\documentclass{article}
\usepackage[preprint]{neurips_2026}
\usepackage{algorithm}
\usepackage{algorithmic}
\usepackage{bm,bbm}
\usepackage{multirow}
\usepackage{mathtools}
\usepackage{amsmath,amsfonts,amssymb,amsthm}
\usepackage{booktabs}
\usepackage{array}
\usepackage{graphicx}
\usepackage{xcolor}
\usepackage{hyperref}
\usepackage{url}
\usepackage{fontawesome5}
\usepackage{subcaption}
\usepackage{colortbl}
\usepackage{float}
\usepackage{placeins}
\usepackage{listings}
\definecolor{respoGray}{gray}{0.88}
\definecolor{tableHeader}{RGB}{235,239,244}
\definecolor{tableAverage}{RGB}{246,247,249}
\definecolor{respoHighlight}{RGB}{255,242,232}
\definecolor{codeBackground}{RGB}{247,248,250}
\definecolor{codeBorder}{RGB}{210,214,220}
\definecolor{codeKeyword}{RGB}{44,82,130}
\definecolor{codeString}{RGB}{153,76,0}
\definecolor{codeComment}{RGB}{88,105,88}
\lstdefinestyle{pythoncode}{
  language=Python,
  basicstyle=\ttfamily\footnotesize,
  keywordstyle=\color{codeKeyword}\bfseries,
  stringstyle=\color{codeString},
  commentstyle=\color{codeComment}\itshape,
  backgroundcolor=\color{codeBackground},
  frame=single,
  rulecolor=\color{codeBorder},
  framerule=0.45pt,
  framesep=5pt,
  xleftmargin=0.4em,
  xrightmargin=0.4em,
  aboveskip=0.65em,
  belowskip=0.65em,
  columns=fullflexible,
  keepspaces=true,
  showstringspaces=false,
  breaklines=true,
  tabsize=4
}

\newcommand{\E}{\mathbb{E}}
\newcommand{\KL}{\mathrm{KL}}
\newcommand{\pitheta}{\pi_\theta}
\newcommand{\piold}{\pi_{\mathrm{old}}}

\newcommand{\respoCell}[1]{\cellcolor{respoHighlight}#1}
\newcommand{\respoMethodRow}{\rowcolor{respoHighlight}\cellcolor{white}}
\newcommand{\accerr}[2]{#1_{\scriptscriptstyle\pm #2}}
\newcommand{\projecturl}{https://yhangchen.github.io/ReSPO}

\title{ReSPO: Reshaped Sequence Policy Optimization for Gradient Starvation in Off-Policy Learning}

\author{
  Yihang Chen\quad Yuanhao Ban\quad Cho-Jui Hsieh\\
  Department of Computer Science, University of California, Los Angeles\\
  \texttt{\{yhangchen, banyh2000, chohsieh\}@cs.ucla.edu}
  \\[0.6em]
  \textbf{Project Page: \href{\projecturl}{\textcolor{purple}{\faIcon{globe}~ReSPO}}}
}

\date{}

\begin{document}

\maketitle

\begin{abstract}
Reinforcement learning from verifiable rewards (RLVR) frequently reuses rollouts across multiple policy updates, increasing the mismatch between the current policy and the data-generating policy. We identify a sign-dependent \emph{gradient starvation} problem in clipped policy optimization: clipping suppresses under-generated positive responses at the low-importance-weight tail while permitting severely over-generated negative responses to dominate the high-weight tail. 
To address this, we propose \textbf{ReSPO} (Reshaped Sequence Policy Optimization), which replaces clipping with a smooth, two-branch sequence-level kernel derived from an $\alpha$-divergence variational objective and an exponential variance-control tilt. The positive branch preserves a nonzero gradient weight for under-generated positive responses, while the negative branch suppresses heavily over-generated negative responses. 
We demonstrate that ReSPO effectively learns from long positive reasoning trajectories during early training, even when accumulated policy drift relegates them to the low-importance-weight tail. On dense and MoE Qwen3 models, ReSPO accelerates early optimization, improves final training scores, and achieves higher held-out benchmark performance under a rollout reuse, validating our approach on importance-weight tail control in off-policy learning. 
\end{abstract}

\section{Introduction}
RL from verifiable rewards (RLVR) improves LLM reasoning using scalar rewards~\citep{chen2025learning,deepseek2025r1,qwen2025qwen3}, and GRPO~\citep{shao2024deepseekmath} computes group-relative advantages without a value function. Expensive and slow generation motivates rollout reuse, making the data \emph{off-policy} in training: the current policy $\pitheta$ differs from the old policy $\piold$ that generated the responses. The sequence importance ratio $W=\pitheta(\bm{o}\mid\bm{q})/\piold(\bm{o}\mid\bm{q})$ measures this mismatch on the prompt $\bm{q}$ and response $\bm{o}$. A small ratio $W$ means that the response is now less likely to be generated by the current policy than the old policy, while a large $W$ means that it is more likely. GRPO clips the token importance ratio $w_t=\pitheta(o_t|\bm{q},\bm{o}_{<t})/\piold(o_t|\bm{q},\bm{o}_{<t})$ and GSPO~\citep{zheng2025gspo} clips the length-normalized ratio $s=W^{1/|\bm{o}|}$ to stabilize training. Both methods use clipping, so their in-region gradients scale \emph{linearly} with the importance ratio.

This linear weighting starves informative gradients in opposite ways depending on the advantage sign. We call responses with $\hat{A}>0$ \emph{positive responses} and those with $\hat{A}<0$ \emph{negative responses}, according to their group-relative reward. For positive responses, the gradient is proportional to $w_t$, so already familiar tokens near the upper clip threshold receive the largest updates, while under-generated tokens with $w_t\ll1$ receive almost none. For negative responses, heavily over-generated tokens with $w_t\gg1$ remain unclipped, so their unbounded weight can overshadow the gradient contributions from other tokens. The same failure carries over to the sequence-level ratio used by GSPO. We refer to this sign-dependent misallocation of learning signal as \emph{gradient starvation}.

We tackle the failures by reshaping the importance weight $W$ at the sequence level. Instead of clipping, we replace the raw $W$ with a smooth, sign-dependent kernel $\phi^{(\pm)}(W)$, and the two failures above determine the shape we need: $\phi^{(+)}(0)>0$ must preserve under-generated positive responses, while $\phi^{(-)}(0)=\phi^{(-)}(\infty)=0$ must suppress both already-suppressed and severely over-generated negative responses. GRPO and GSPO clipping miss these critical tails: their effective weight vanishes on the low-ratio tail but grows without bound on the high-ratio tail. VESPO~\citep{shen2025vespo} suppresses both tails, fitting negative advantages but starving positive ones at $W\to0$. We propose \textbf{ReSPO} (Reshaped Sequence Policy Optimization), which derives a smooth two-branch kernel from an $\alpha$-divergence variational objective with exponential tilting for variance control. The choice $\alpha^{(+)}>1$ gives the required positive tail, while the $\alpha^{(-)}=1$ limit gives both required negative tails. On DAPO-MATH~\citep{yu2025dapo} with Qwen3-1.7B-Base and Qwen3-30B-A3B-Base, ReSPO attains a higher late-stage training score than every clipped baseline in all six model--$N$ settings and a higher score than VESPO in five of six.

\section{Related Work}

\textbf{Policy optimization for LLM reasoning.}
PPO~\citep{schulman2017ppo} is the standard method for RLHF~\citep{ouyang2022training}, using a clipped surrogate for stable updates. With automated verifiers~\citep{deepseek2025r1}, value-free alternatives have become popular. GRPO~\citep{shao2024deepseekmath} computes group-relative advantages and clips per-token ratios; DAPO~\citep{yu2025dapo} adds decoupled clipping and dynamic sampling; and REINFORCE++~\citep{hu2025reinforce} uses global advantage normalization. These methods focus on token-level updates, whereas we study sequence-level gradient starvation. REAL~\citep{zhai2025rewards} reframes gradient starvation as a classification problem; we retain importance sampling to keep off-policy staleness explicit. This distinction matters increasingly as rollout reuse makes each batch more off-policy over successive optimizer steps.

\textbf{Sequence-level and off-policy optimization.}
Token-level importance correction is mismatched with sequence-level rewards. GSPO~\citep{zheng2025gspo} uses a length-normalized sequence ratio for stable MoE training, while VESPO~\citep{shen2025vespo} derives $\phi_{\KL}(W) = W^\beta \exp(\lambda(1 - W))$ from a KL variational objective. Reusing rollouts across mini-batches causes off-policy drift~\citep{noukhovitch2025async}, especially for MoE models~\citep{zheng2025gspo}; Routing Replay~\citep{ma2025routing} and truncated importance sampling~\citep{liu2025speed} have been proposed to improve system stability. ReSPO complements these techniques by controlling how the importance weight allocates gradient across responses. We generalize VESPO's framework to the $\alpha$-divergence family with a sign-dependent two-branch kernel. Sec.~\ref{sec:two_branch} and Fig.~\ref{fig:phi_comparison} compare these kernel designs.

\textbf{Trust regions and divergence measures.}
TRPO~\citep{schulman2015trpo} limits updates with a KL bound, and PPO approximates this with hard clipping. Softer options include probability smoothing~\citep{dwyer2025soft} and adaptive gating~\citep{gao2025sapo}. R\'enyi-divergence bounds have also been used to analyze the variance of importance-sampling estimators in policy optimization~\citep{metelli2018policy}. ReSPO instead uses separable power-Bregman geometry to derive a sequence-level reshaping family. Its two branches allocate gradient according to the advantage sign and the importance-weight tail behavior that matters for off-policy optimization.

\section{Method}

\subsection{Preliminaries: Token-Level and Sequence Policy Optimization}
\label{sec:preliminaries}

\textbf{LLM and GRPO.}
A large language model (LLM) parameterized by $\theta$ defines an autoregressive policy $\pitheta(\bm{o} \mid \bm{q}) = \prod_{t=1}^{T} \pitheta(o_t \mid \bm{q}, o_{<t})$ that generates a response $\bm{o} = (o_1, \ldots, o_T)$ for a prompt $\bm{q}$. In the RLVR setting, a deterministic scalar reward $R(\bm{q}, \bm{o})$ gives a sequence-level signal. GRPO samples $G$ responses $\{\bm{o}_i\}_{i=1}^G$ from the old policy $\piold$ and forms the group-normalized advantage $\hat{A}_i = (R(\bm{q}, \bm{o}_i) - \mathrm{mean}_j R(\bm{q}, \bm{o}_j)) /(\mathrm{std}_j R(\bm{q}, \bm{o}_j) + \epsilon_A)$; implementation details are given in App.~\ref{app:hyperparams}. GRPO then optimizes a token-level clipped surrogate:
\begin{equation}\label{eq:grpo}
  \mathcal{J}_{\mathrm{GRPO}}(\theta) = \E_{\bm{q}, \{\bm{o}_i\}} \left[ \frac{1}{G} \sum_{i=1}^{G} \frac{1}{|\bm{o}_i|} \sum_{t=1}^{|\bm{o}_i|} \min\!\Big(w_{i,t}\, \hat{A}_i,\; \mathrm{clip}(w_{i,t}, 1{-}\varepsilon, 1{+}\varepsilon)\, \hat{A}_i \Big) \right],
\end{equation}
where $w_{i,t} = \pitheta(o_{i,t} \mid \bm{q}, \bm{o}_{i,<t}) / \piold(o_{i,t} \mid \bm{q}, \bm{o}_{i,<t})$ is the per-token importance ratio. Because the reward is defined at the sequence level, applying a separate single-sample ratio to each token does not recover the exact sequence-level importance correction and can introduce noisy token-wise scaling as sequence length grows~\citep{shen2025vespo,zheng2025gspo}. Routing fluctuations create an additional source of training--inference mismatch in MoE models, motivating stabilization techniques such as Routing Replay~\citep{ma2025routing,zheng2025gspo}. These instabilities compound as policy staleness grows.

\textbf{Sequence-level ratio: GSPO.}
A natural fix is to work on the sequence-level importance weight $W(\bm{o}) = \prod_t w_t = \pitheta(\bm{o} \mid \bm{q}) / \piold(\bm{o} \mid \bm{q})$, which aggregates token ratios at the same granularity as the reward. GSPO~\citep{zheng2025gspo} puts this view into practice with a length-normalized geometric-mean ratio $s_i(\theta) = (\pitheta(\bm{o}_i \mid \bm{q}) / \piold(\bm{o}_i \mid \bm{q}))^{1/|\bm{o}_i|}$ and a clipped surrogate. 
\begin{equation}\label{eq:gspo}
  \mathcal{J}_{\mathrm{GSPO}}(\theta) = \E_{\bm{q},\{\bm{o}_i\}}\!\left[ \frac{1}{G}\sum_{i=1}^{G} \min\!\Big( s_i(\theta)\,\hat{A}_i,\; \mathrm{clip}(s_i(\theta), 1{-}\varepsilon_{\mathrm{low}}, 1{+}\varepsilon_{\mathrm{high}})\,\hat{A}_i \Big) \right].
\end{equation}
GSPO trains more stably than GRPO, notably without Routing Replay on MoE models~\citep{zheng2025gspo}, so the sequence-level weight is a natural starting point for off-policy reshaping. However, hard clipping of $s_i$ keeps GRPO's gradient starvation inside the unclipped region. Moreover, the normalization $W^{1/|\bm{o}_i|}$ induces length-dependent bias because it tempers the sequence importance ratio by an exponent that depends on response length (see App.~\ref{app:length_bias}).

\subsection{Gradient Starvation}
\label{sec:gradient_misallocation}

Let $\rho=w_t$ in GRPO and $\rho=s_i$ in GSPO. We call a response positive when the estimated advantage $\hat{A}>0$ and negative when $\hat{A}<0$.
Inside the unclipped region, both objectives weight the gradient linearly by $\rho$. For GRPO and GSPO, the gradients are
\begin{equation*}
  \nabla_\theta \mathcal{J}_{\mathrm{GRPO}} \propto \hat{A}  w_t  \nabla_\theta \log \pitheta(o_t \mid \bm{q}, \bm{o}_{<t}),\ \ \text{and}\ \  \nabla_\theta \mathcal{J}_{\mathrm{GSPO}}
  \propto  \frac{\hat{A}_i s_i}{|\bm{o}_i|}\sum_{t=1}^{|\bm{o}_i|}
     \nabla_\theta\log\pitheta(o_{i,t}\mid\bm{q},\bm{o}_{i,<t}).
\end{equation*}
Apart from the length factor in GSPO, the gradient magnitude of either objective grows linearly with $|\hat{A}|\rho$. Clipping discards large-ratio positive and small-ratio negative updates, but leaves two failures~\citep{zhai2025rewards}:
\begin{itemize}
  \item \textbf{Starved positive update.} When $\hat{A}>0$ and $\rho\to0$, a positive response is becoming less likely, but its recovery signal vanishes: $w_t=0.01$ receives $100\times$ less weight than $w_t=1$.
  \item \textbf{Dominating negative update.} When $\hat{A}<0$ and $\rho\to\infty$, a negative response is becoming more likely. This side remains unclipped, so its weight grows without bound and can dominate repeated off-policy updates.
\end{itemize}

ReSPO addresses these GRPO failures by reshaping $W$. The likelihood shift encoded by $W$, together with the advantage sign, gives the four cases below.

\begin{center}
  \textbf{How ReSPO should treat each tail. $\uparrow$ means ``reinforce'' and $\downarrow$ means ``suppress''}\\[3pt]
  \small
  \setlength{\tabcolsep}{3pt}
  \renewcommand{\arraystretch}{1.12}
  \begin{tabular}{@{}>{\centering\arraybackslash}p{0.1\linewidth}>{\raggedright\arraybackslash}p{0.40\linewidth}>{\raggedright\arraybackslash}p{0.40\linewidth}@{}}
    \toprule
    \rowcolor{tableHeader}
    \textbf{Response} & \textbf{$W\to0$: less likely now} & \textbf{$W\to\infty$: more likely now} \\
    \midrule
     $\hat{A}>0$ & \textbf{$\uparrow$:} positive response being lost & \textbf{$\downarrow$:} positive response already reinforced \\
     $\hat{A}<0$ & \textbf{$\downarrow$:} negative response already suppressed & \textbf{$\downarrow$:} negative response being amplified \\
    \bottomrule
  \end{tabular}
\end{center}

Therefore, ReSPO requires $\phi^{(+)}(0)>0$ and $\phi^{(+)}(\infty)=\phi^{(-)}(0)=\phi^{(-)}(\infty)=0$.

\subsection{ReSPO: Reshaped Sequence Policy Optimization}
\label{sec:respo}

To address gradient starvation, ReSPO applies a soft kernel to the sequence-level importance weight. The derivation has three stages: (i) interpret reshaping as a change of measure; (ii) use an $\alpha$-divergence objective to obtain an unconstrained kernel $\phi_0$; and (iii) project this measure under a moment constraint, producing a smooth exponential tilt $\phi$. We then choose $\alpha$ by advantage sign so that the two branches satisfy the requirements above; App.~\ref{app:derivations} provides the full derivations.

\subsubsection{Weight Reshaping as Measure Change}
\label{sec:measure_change}

To find a $\phi$ that satisfies the properties stated in Sec.~\ref{sec:gradient_misallocation}, we first ask what replacing the raw importance weight $W$ with a generic reshaped weight $\phi(W)$ actually does to the gradient. Let $\mu = \piold$ denote the behavior policy from which rollouts are sampled and $\pi = \pitheta$ the current target policy, so that the sequence-level importance weight defined in Sec.~\ref{sec:preliminaries} is $W(\bm{o}) = \pi(\bm{o} \mid \bm{q}) / \mu(\bm{o} \mid \bm{q})$. For the reward function $R(\cdot)$, the policy gradient of $\mathbb{E}_{\bm{o}\sim\pi}R(\bm{o})$ is
\begin{equation}
  \nabla_\theta\mathbb{E}_{\bm{o}\sim\pi} R(\bm{o}) = \mathbb{E}_{\bm{o}\sim\pi} R(\bm{o}) \nabla_\theta\log\pi_\theta(\bm{o})=\mathbb{E}_{\bm{o}\sim\mu} W(\bm{o})\cdot R(\bm{o})\cdot \nabla_\theta\log\pi_\theta(\bm{o})\,.
\end{equation}
Applying any reshaping function $\phi: \mathbb{R}_+ \to \mathbb{R}_+$ to $W$ implicitly defines an unnormalized measure $\tau$. The measure-change identity is
\begin{equation}\label{eq:measure_change}
  \E_{\bm{o} \sim \mu}\!\big[\phi(W(\bm{o})) \cdot R(\bm{o})\cdot \nabla_\theta\log \pitheta(\bm{o})\big] = \sum_{\bm{o}}\tau(\bm{o})\,R(\bm{o})\cdot \nabla_\theta\log \pitheta(\bm{o}),
\end{equation}
where $\tau$ is given by
\begin{equation}\label{eq:proposal}
  \tau(\bm{o}) = \mu(\bm{o})\, \phi(W(\bm{o})).
\end{equation}
Eq.~\eqref{eq:proposal} fixes the notation used below: $W$ is the raw sequence ratio, $\phi$ is its reshaping kernel, and $\tau=\mu\phi$ is an unnormalized measure. Let $Z := \sum_{\bm{o}}\tau(\bm{o}) = \E_{\mu}[\phi(W)]$ and $\bar{\tau}(\bm{o}) := \tau(\bm{o})/Z$. Then only $\bar{\tau}$ is a probability distribution, and $\sum_{\bm{o}}\tau(\bm{o})g(\bm{o}) = Z\,\E_{\bm{o}\sim\bar{\tau}}[g]$.
This reframing suggests a design strategy. Rather than hand-picking $\phi$ as clipping does, we first solve for an unnormalized $\tau$, read off $\phi$ through Eq.~\eqref{eq:proposal}, and finally verify its tails against the requirements in Sec.~\ref{sec:gradient_misallocation}.

\subsubsection{Variational Derivation of the Reshaping Kernel}
\label{sec:variational}

Our goal is to design a measure $\tau$ that balances three goals: staying close to the behavior distribution $\mu$ for sample efficiency, moving toward the target policy $\pi$ to reduce bias, and controlling variance for training stability. We begin with a brief review of the Bregman divergence.

\textbf{Bregman divergence and the dual-proximity objective.}
Let $f: \mathbb{R}_+ \to \mathbb{R}$ be a strictly convex function, which we call the Bregman generator. The Bregman divergence induced by $f$ for (unnormalized) measures $p,q\geq 0$ is
\begin{equation}\label{eq:bregman}
  D_f(p \,\|\, q) = \sum_{\bm{o}} \Big[ f\big(p(\bm{o})\big) - f\big(q(\bm{o})\big) - f'\big(q(\bm{o})\big)\,\big(p(\bm{o}) - q(\bm{o})\big) \Big].
\end{equation}
Borrowing the idea from annealed importance sampling~\citep{neal2001ais}, we formulate the dual-proximity objective for a general Bregman divergence:
\begin{equation}\label{eq:variational}
  \min_{\tau \geq 0} \;\; (1-\beta)\, D_f(\tau \,\|\, \mu) \;+\; \beta\, D_f(\tau \,\|\, \pi),
\end{equation}
where $\beta \in (0, 1)$ controls the trade-off. At $\beta = 0$ we aim to recover the behavior policy ($\tau = \mu$, zero importance-weight variance but maximum bias), and at $\beta = 1$ we target the current policy ($\tau = \pi$, unbiased but with possibly high variance). Throughout this paper, $\tau$ is an unnormalized measure; only $\bar{\tau}$ denotes its normalized probability distribution. See App.~\ref{app:renyi_fo} for discussion.

\textbf{Choice of Bregman generator.}
The choice of $f$ determines the geometry of the interpolation and, with it, the form and tail behavior of the reshaping kernel.

\textbf{KL divergence ($f(x) = x \log x - x$).}\; This is the choice in VESPO~\citep{shen2025vespo}. The first-order condition of Eq.~\eqref{eq:variational} gives the geometric mixture $\tau^*_{\KL}(\bm{o}) = \mu(\bm{o})^{1-\beta} \cdot \pi(\bm{o})^{\beta}$, with reshaping kernel $\phi_{\KL}(W) = W^{\beta}$. This is also the continuous $\alpha\to1$ limit of the power-mean family below. After exponential tilting it provides the two vanishing tails required by the negative branch, but its value at $W\to0$ cannot satisfy the positive branch.

\textbf{$\alpha$-divergence ($f(x) = \frac{1}{\alpha(\alpha-1)} x^{\alpha}$, $\alpha \neq 0, 1$).}\; $f$ is strictly convex on $\mathbb{R}_+$ for all $\alpha \notin \{0,1\}$. Solving the first-order condition gives the power-mean interpolation
\begin{equation}\label{eq:tau_renyi}
  \tau^*_{0}(\bm{o}) = \Big[ (1-\beta)\, \mu(\bm{o})^{\alpha-1} + \beta\, \pi(\bm{o})^{\alpha-1} \Big]^{\frac{1}{\alpha-1}},
\end{equation}
and substituting $\pi(\bm{o}) = \mu(\bm{o})\, W(\bm{o})$ together with Eq.~\eqref{eq:proposal} yields the corresponding unconstrained reshaping kernel
\begin{equation}\label{eq:phi_unconstrained}
  \phi_0(W) = \big[ (1-\beta) + \beta\, W^{\alpha-1} \big]^{1/(\alpha-1)}.
\end{equation}
This kernel is valid for both $0 < \alpha < 1$ and $\alpha > 1$ and satisfies $\phi_0(1) = 1$ by construction. The full first-order derivation is given in App.~\ref{app:renyi_fo} for completeness.

\subsubsection{Exponential Tilt for Variance Control}
\label{sec:exp_tilt}

The unconstrained kernel $\phi_0$ from Eq.~\eqref{eq:phi_unconstrained} redistributes gradient weight and requires a separate variance-control step. Since $\tau=\mu\,\phi$, the natural second-moment target is $\E_\mu[\phi(W)^2]=\sum_{\bm{o}}\tau(\bm{o})\phi(W(\bm{o}))$. The final $\phi$ is still unknown, so we replace it inside the constraint by the fixed proxy $\phi_0$ and require $\sum_{\bm{o}}\tau(\bm{o})\phi_0(W(\bm{o}))\leq C_1$. This proxy is sufficient because the resulting tilt leads to bounded $\E_\mu[\phi(W)^2]$, as shown in App.~\ref{app:pos_lagrangian}. We additionally impose $\sum_{\bm{o}}\tau(\bm{o})\leq C_2$ to prevent the measure from having unbounded mass.

For this projection step, we use the generalized KL divergence for unnormalized measures $p,q\geq 0$: $D_{\text{KL}}(p\|q)
  = \sum_{\bm{o}} \left( p(\bm{o}) \log \frac{p(\bm{o})}{q(\bm{o})} - p(\bm{o}) + q(\bm{o}) \right)$.
We find the closest measure to $\tau_0^*$ that satisfies both constraints in this KL geometry~\citep{csiszar1975divergence}:
\begin{equation}\label{eq:kl_proj}
  \tau^* = \arg\min_{\tau}\; D_{\KL}(\tau \,\|\, \tau_0^*) \quad \text{s.t.} \quad \sum_{\bm{o}}\tau(\bm{o})\phi_0(W(\bm{o})) \leq C_1, \quad \sum_{\bm{o}}\tau(\bm{o})\leq C_2.
\end{equation}
The KKT conditions for this KL projection give an exponential tilt of $\tau_0^*$ with sufficient statistic $\phi_0(W)$:
\begin{equation}\label{eq:exp_tilt}
  \tau^*(\bm{o}) \propto \tau_0^*(\bm{o}) \cdot \exp\!\big(-\lambda\, \phi_0(W(\bm{o}))\big),
\end{equation}
where $\lambda \geq 0$ is the Lagrange multiplier. Extracting the reshaping kernel via $\tau=\mu\,\phi(W)$ and using $\phi_0(1)=1$ to normalize $\phi(1)=1$ gives
\begin{equation}\label{eq:phi_exp}
  \phi(W) = \underbrace{\phi_0(W)}_{\text{$\alpha$ reshaping}} \cdot \underbrace{\exp\!\big(\lambda\,(1 - \phi_0(W))\big)}_{\text{exponential tilt}}.
\end{equation}
The two divergences therefore play different roles: the $\alpha$-divergence determines the power-mean interpolation and its tails, while the KL projection turns a linear moment constraint into a smooth multiplicative tilt rather than a hard boundary (App.~\ref{app:pos_lagrangian}). As $\alpha \to 1^+$, $\phi_0(W) \to W^{\beta}$ (App.~\ref{app:vespo_limit}) and the kernel becomes $W^{\beta}\exp(\lambda(1-W^{\beta}))$. This differs from VESPO~\citep{shen2025vespo}, whose exponential factor acts on the raw weight $W$ rather than the reshaped weight $W^\beta$.

\subsubsection{Tail Behavior and the Two-Branch Design}
\label{sec:two_branch}

The kernel of Eq.~\eqref{eq:phi_exp} depends on three hyperparameters $(\alpha, \beta, \lambda)$, and its tail behavior determines whether the gradient requirements of Sec.~\ref{sec:gradient_misallocation} are met. We extend the power-mean kernel continuously to $\alpha=1$ by defining $\phi_0(W)=W^\beta$ (App.~\ref{app:vespo_limit}). The key observation is that no single value of $\alpha$ satisfies both the positive- and negative-advantage requirements simultaneously.

\textbf{Tail analysis.}
For $\lambda>0$, the tilt sends $\phi(W)$ to zero whenever $\phi_0(W)$ diverges. For $\alpha>1$, $\phi(0)>0$ and $\phi(\infty)=0$, matching the positive-branch requirements. For $0<\alpha<1$, $\phi(0)=0$ but $\phi(\infty)>0$, so extremely over-generated negative responses retain nonzero weight, and these outliers could impact the training. At the boundary $\alpha=1$, $\phi(0)=\phi(\infty)=0$, exactly matching the negative-branch requirements. This is where the sign-specific design enters the variational construction.

\textbf{The ReSPO kernel.}
We use a sign-dependent parameterization: $\alpha^{(+)} > 1$ for positive advantages and the $\alpha^{(-)}=1$ limiting form for negative advantages, both with $0<\beta^{(\pm)}<1$. The full ReSPO reshaping kernel is
\begin{equation}\label{eq:phi_two_branch} \phi(W; \hat{A}) = 
 \phi_0^{(\pm)}(W) \cdot \exp\!\big(\lambda^{(\pm)}\,(1-\phi_0^{(\pm)}(W))\big),
\end{equation}
where $(+)$ applies when $\hat{A}\geq0$ and $(-)$ applies when $\hat{A}<0$, and
\begin{equation}\label{eq:phi0_two_branch}
  \phi_0^{(+)}(W) = \left[(1-\beta^{(+)}) + \beta^{(+)}W^{\alpha^{(+)}-1}\right]^{1/(\alpha^{(+)}-1)},
  \qquad
  \phi_0^{(-)}(W) = W^{\beta^{(-)}}.
\end{equation}
The negative branch is the $\alpha^{(-)}\to1$ limit of the same family. Both branches satisfy $\phi(1)=1$, and together they realize all four sign-specific tail requirements. Combining the kernel with group-normalized advantages gives the ReSPO policy gradient for a sampled batch from Eq.~\eqref{eq:measure_change}:
\begin{equation}\label{eq:respo}
\begin{aligned}
  \nabla_\theta \mathcal{J}_{\mathrm{ReSPO}} = \E_{\bm{q},\, \{\bm{o}_i\} \sim \piold} \left[ \frac{1}{G} \sum_{i=1}^{G} {\rm sg}\left(\phi(W_i; \hat{A}_i)\right) \cdot \hat{A}_i \cdot\sum_{t=1}^{|\bm{o}_i|} \nabla_\theta \log \pitheta(o_{i,t} \mid \bm{q}, \bm{o}_{i,<t}) \right],
\end{aligned}
\end{equation}
Here ${\rm sg}(\cdot)$ is the stop-gradient operator and $W_i = \prod_{t=1}^{|\bm{o}_i|} w_{i,t}$ is the sequence-level importance weight for response $i$.
We apply the kernel to the unnormalized ratio $W$, because the variational derivation (Sec.~\ref{sec:variational}) operates on distributions over sequences and $W$ is the natural Radon--Nikodym derivative; length normalization in GSPO would distort the geometry. The detached weight $\phi(W_i; \hat{A}_i)$ acts as a per-sequence gradient scaling coefficient in a REINFORCE-style estimator. For an on-policy update, $W=1$ and both branches reduce to unit weight (App.~\ref{app:on_policy_limit}). App.~\ref{app:implementation} gives the numerically stable log-space form, and Alg.~\ref{alg:respo} summarizes the complete update.

\subsubsection{Hyperparameter Selection}
\label{sec:hyperparams}

The three parameters have separate roles: $\alpha$ determines the tail regime, $\beta$ sets the behavior--target interpolation, and $\lambda$ controls the exponential tilt. The tail analysis fixes $\alpha$ by advantage sign. We set $\beta^{(\pm)}=0.5$ for a symmetric interpolation, then choose $\lambda^{(\pm)}$ according to the desired local shape.

\textbf{Positive branch: $\alpha^{(+)} = 2$, $\beta^{(+)} = 0.5$, $\lambda^{(+)} = 2$.}
At $\alpha^{(+)}=2$, the power mean reduces to the linear form $\phi_0^{(+)}(W)=(1-\beta^{(+)})+\beta^{(+)}W$. Setting $\beta^{(+)}=0.5$ gives the arithmetic mean $\phi_0^{(+)}(W)=\tfrac{1}{2}(1+W)$, treating $\mu$ and $\pi$ symmetrically. We require $\phi^{(+)}{}'(0)=0$, which gives $\lambda^{(+)}=1/(1-\beta^{(+)})=2$ and makes $W=0$ the global maximum. Although an importance weight cannot be negative, the linear closed form extends to the real line and has its unique maximum at zero (App.~\ref{app:hyperparam_derivations}). The resulting kernel is
$\phi^{(+)}(W)=\frac{1+W}{2}\exp(1-W)$,
which decreases strictly for $W>0$ and has maximum $\phi^{(+)}(0)=\tfrac{1}{2}e\approx1.36$.

\begin{figure}[H]
  \centering
  \begin{minipage}[c]{0.69\linewidth}
    \centering
    \includegraphics[width=\linewidth]{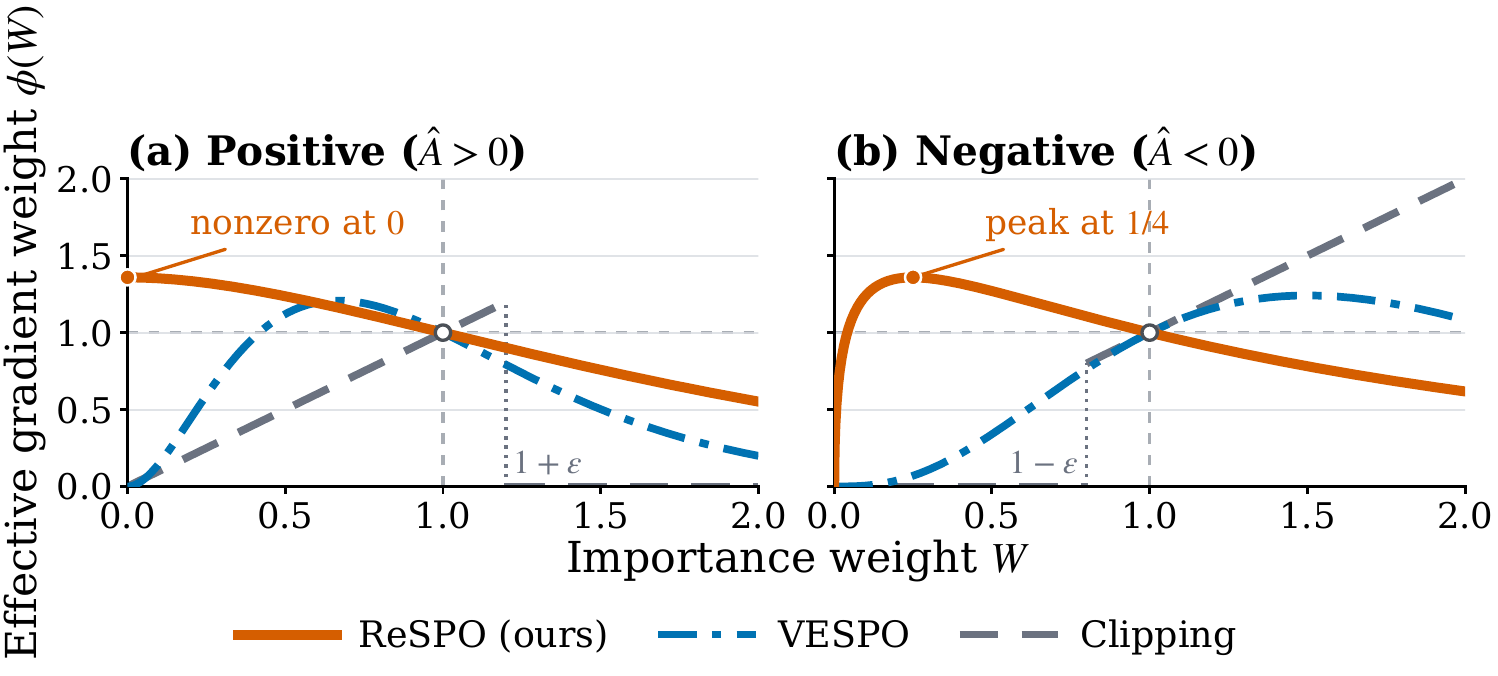}
  \end{minipage}\hfill
  \begin{minipage}[c]{0.295\linewidth}
    \centering
    \normalsize
    \textbf{Tail behavior}\par\vspace{4pt}
    \setlength{\tabcolsep}{2pt}
    \renewcommand{\arraystretch}{1.18}
    \begin{tabular}{@{}lcc@{}}
      \toprule
      \rowcolor{tableHeader}
      \textbf{Method} & \shortstack{$\phi^{(+)}$\\$(0,\infty)$} & \shortstack{$\phi^{(-)}$\\$(0,\infty)$} \\
      \midrule
      Clipping & $(0,0)$ & $(0,\infty)$ \\
      VESPO & $(0,0)$ & $(0,0)$ \\
      \respoCell{\textbf{ReSPO}} & \respoCell{$(e/2,0)$} & \respoCell{$(0,0)$} \\
      \bottomrule
    \end{tabular}
  \end{minipage}
  \caption{Kernel shapes and tail limits ($\varepsilon=0.2$). The adjacent table gives the limits as $W\to(0,\infty)$. ReSPO preserves positive-branch signal near zero while suppressing both negative tails.}
  \label{fig:phi_comparison}
\end{figure}

\textbf{Negative branch: $\alpha^{(-)} = 1$, $\beta^{(-)} = 0.5$, $\lambda^{(-)} = 2$.}
The choice $\alpha^{(-)}=1$ is made so that $\phi^{(-)}(0)=\phi^{(-)}(\infty)=0$. With $\beta^{(-)}=0.5$, its unconstrained kernel is the geometric mean $\phi_0^{(-)}(W)=\sqrt{W}$. We choose $\lambda^{(-)}$ by matching the two branches at the on-policy point: $\phi^{(+)}{}'(1)=-\tfrac{1}{2}$, while $\phi^{(-)}{}'(1)=\beta^{(-)}(1-\lambda^{(-)})$. Equating the derivatives gives $\lambda^{(-)}=2$ (App.~\ref{app:hyperparam_derivations}) and
$\phi^{(-)}(W)=\sqrt{W}\exp\!\left(2(1-\sqrt{W})\right)$.
Thus the branches match in value and slope at $W=1$. Fig.~\ref{fig:phi_comparison} shows that the negative kernel peaks at $W^*=\frac{1}{4}$ with value $e/2$ and vanishes on both tails.

\section{Experiments}
\label{sec:experiments}

We evaluate ReSPO on mathematical reasoning with one dense model, \textbf{Qwen3-1.7B-Base}, and one mixture-of-experts (MoE) model, \textbf{Qwen3-30B-A3B-Base}. For each model we vary the rollout-reuse ratio $N$ to study how the methods degrade as off-policy drift grows.

\subsection{Experimental Setup}
\label{sec:experimental_setup}

\textbf{Training \& evaluation protocol.}
All experiments use \texttt{verl}~\citep{sheng2025verl}, asynchronous vLLM rollouts~\citep{kwon2023pagedattention}, mini-batch size $M=32$, $G=8$ rollouts per prompt, and the GRPO advantage estimator. The dense model uses FSDP with AdamW, whereas the MoE model uses Megatron with Adam; both use a learning rate of $10^{-6}$ and no KL penalty. Tab.~\ref{tab:hyperparams_training} lists the shared training settings. Both models are trained on DAPO-MATH-17k~\citep{yu2025dapo}; we use a strict-box verifier as the training reward (App.~\ref{app:dapo_reward}). The prompt limit is $1{,}024$ tokens, and the response limits are $15{,}360$ and $8{,}192$ for Qwen3-1.7B and Qwen3-30B-A3B. The 30B setting adds a soft length penalty: it is zero through $4{,}096$ tokens and decreases linearly to $-1$ at the $8{,}192$-token limit, encouraging concise responses without truncating them at $4{,}096$. We set the global batch size to $NM$ and split each rollout batch into $N$ consecutive mini-batches. Thus larger $N$ reuses the same $\piold$ for more updates and induces more off-policy drift. We evaluate $N\in\{8,16,32\}$ and fix every run to $1024$ policy updates, corresponding to $32{,}768$ training prompts. Our evaluation uses the full test splits of AIME 2025, AIME 2024, and AMC 2023~\citep{aime2025,aime2024,amc2023}, together with OlympiadBench~\citep{he2024olympiadbench}, MinervaMath~\citep{lewkowycz2022minerva}, and MATH-500~\citep{hendrycks2021math,lightman2024let}. We sample with temperature $1.0$, top-$p=0.7$, and no top-$k$ truncation, allowing a maximum response length of $16{,}384$. For each problem, we generate $16$ responses on AIME 2025, AIME 2024, and AMC 2023, and $4$ on each remaining benchmark; each completion is scored using Math-Verify~\citep{kydlicek2024mathverify} (App.~\ref{app:validation_verifier}).

\textbf{Baselines, metrics, and reporting.}
We compare with GRPO~\citep{shao2024deepseekmath}, GSPO~\citep{zheng2025gspo}, and VESPO~\citep{shen2025vespo} under matched data and optimizer settings. Tabs.~\ref{tab:hyperparams_training} and~\ref{tab:hyperparams_method} list the shared and method-specific settings, respectively. In particular, we use VESPO's published best-performing sign-specific setting, $(\beta^{(+)},\lambda^{(+)})=(2,3)$ and $(\beta^{(-)},\lambda^{(-)})=(3,2)$. Tab.~\ref{tab:eval_results} reports per-benchmark results at $N\in\{8,16,32\}$. For training curves, we apply the affine transformation $(\texttt{score}+1)/2$ to the recorded DAPO-MATH score and plot it against the number of policy updates, making experiments with different $N$ directly comparable (App.~\ref{app:dapo_reward}). We call this quantity training accuracy for 1.7B and penalized training accuracy for 30B. We report the maximum logged score within the first $256$ policy updates without an error bar and the policy-step-weighted trapezoidal mean over the last $128$ policy updates with its within-run temporal standard deviation. Tab.~\ref{tab:eval_results} uses a cluster-bootstrap standard error over evaluation problems (App.~\ref{app:bootstrap_uncertainty}). We discuss cross-seed standard deviations in Sec.~\ref{sec:discussion}.

\subsection{Training and Evaluation Performance}
Fig.~\ref{fig:training_accuracy} and Tab.~\ref{tab:eval_results} compare optimization and generalization across both models and all three values of $N$. Figs.~\ref{fig:stats_qwen17b_alpha1} and~\ref{fig:stats_qwen30b_alpha1} additionally report evaluation accuracy, response length, approximate KL divergence, and entropy throughout training.
Normalized training score and held-out pass@1 show the same broad pattern: the observed advantage of ReSPO is generally largest at $N=32$.
 
\textbf{Training performance.}
ReSPO improves quickly and remains higher through most of the training trajectory. It has the highest observed early peak in five of six settings; for 30B at $N=8$, its peak is $1.0$ pp below GRPO. For $N\in\{16,32\}$, its early-peak differences from the highest baseline are $1.0$ and $3.0$ pp on 1.7B and $4.2$ and $5.4$ pp on 30B. ReSPO also has the highest observed last-$128$ mean in five of six settings; on 1.7B at $N=16$, it is $0.4$ pp below VESPO, and each central value lies within the other's temporal range. Its late mean exceeds the highest clipped baseline in all six settings by $4.2$--$8.4$ pp, with non-overlapping temporal ranges. At $N=32$, these differences reach $7.9$ pp on 1.7B and $8.4$ pp on 30B. App.~\ref{app:detailed_comparisons} gives the complete model- and $N$-specific comparison.

\begingroup
\captionsetup{hypcap=false}
\begin{figure}[!htbp]
\centering
\captionsetup{skip=4pt}
\begin{minipage}[c]{0.59\textwidth}
\centering
\includegraphics[width=\linewidth]{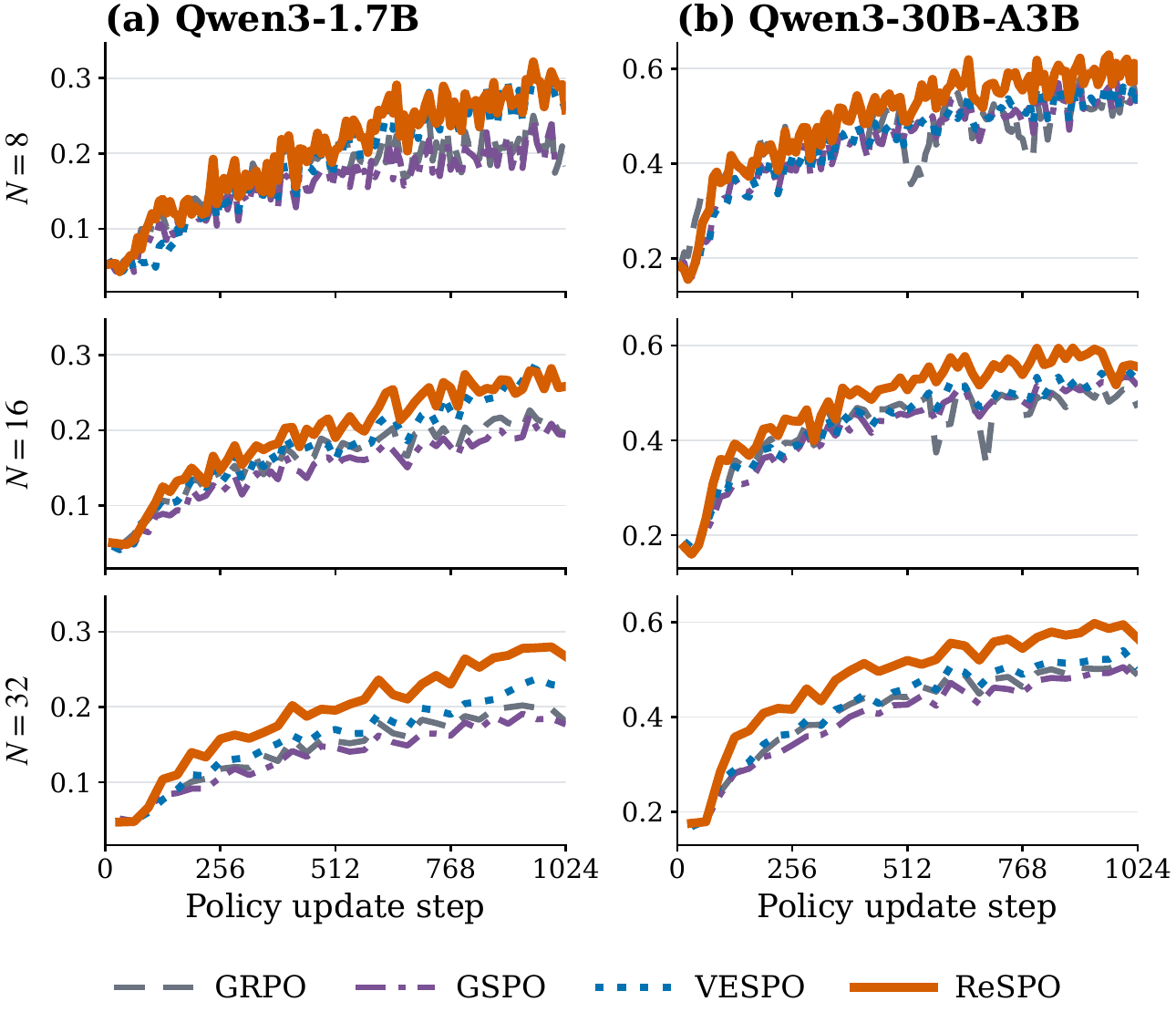}
\end{minipage}\hfill
\begin{minipage}[c]{0.40\textwidth}
\centering
\fontsize{7}{7.4}\selectfont
\newcommand{\figstatnum}[1]{\scalebox{0.80}{$#1$}}
\setlength{\tabcolsep}{0.45pt}
\renewcommand{\arraystretch}{0.94}
\textbf{Early peak ($t\leq256$)}\par\vspace{1.5pt}
\begin{tabular*}{\linewidth}{@{}c@{\extracolsep{\fill}}rrr>{\columncolor{respoHighlight}}r@{}}
\toprule
\rowcolor{tableHeader}
$N$ & \textsc{GRPO} & \textsc{GSPO} & \textsc{VESPO} & \textbf{ReSPO} \\
\midrule
\multicolumn{5}{@{}l}{\textbf{Qwen3-1.7B}} \\
$8$  & \figstatnum{0.169} & \figstatnum{0.158} & \figstatnum{0.161} & \figstatnum{\mathbf{0.192}} \\
$16$ & \figstatnum{0.156} & \figstatnum{0.127} & \figstatnum{0.151} & \figstatnum{\mathbf{0.166}} \\
$32$ & \figstatnum{0.118} & \figstatnum{0.106} & \figstatnum{0.127} & \figstatnum{\mathbf{0.157}} \\
\midrule
\multicolumn{5}{@{}l}{\textbf{Qwen3-30B-A3B}} \\
$8$  & \figstatnum{\mathbf{0.478}} & \figstatnum{0.416} & \figstatnum{0.414} & \figstatnum{0.468} \\
$16$ & \figstatnum{0.403} & \figstatnum{0.369} & \figstatnum{0.386} & \figstatnum{\mathbf{0.445}} \\
$32$ & \figstatnum{0.362} & \figstatnum{0.341} & \figstatnum{0.364} & \figstatnum{\mathbf{0.418}} \\
\bottomrule
\end{tabular*}
\par\vspace{4pt}
\textbf{Late mean (last 128 steps)}\par\vspace{1.5pt}
\begin{tabular*}{\linewidth}{@{}c@{\extracolsep{\fill}}rrr>{\columncolor{respoHighlight}}r@{}}
\toprule
\rowcolor{tableHeader}
$N$ & \textsc{GRPO} & \textsc{GSPO} & \textsc{VESPO} & \textbf{ReSPO} \\
\midrule
\multicolumn{5}{@{}l}{\textbf{Qwen3-1.7B}} \\
$8$  & \figstatnum{\accerr{0.213}{0.017}} & \figstatnum{\accerr{0.209}{0.015}} & \figstatnum{\accerr{\mathbf{0.275}}{0.016}} & \figstatnum{\accerr{\mathbf{0.286}}{0.016}} \\
$16$ & \figstatnum{\accerr{0.210}{0.007}} & \figstatnum{\accerr{0.199}{0.009}} & \figstatnum{\accerr{\mathbf{0.268}}{0.009}} & \figstatnum{\accerr{\mathbf{0.264}}{0.010}} \\
$32$ & \figstatnum{\accerr{0.197}{0.005}} & \figstatnum{\accerr{0.184}{0.003}} & \figstatnum{\accerr{0.230}{0.004}} & \figstatnum{\accerr{\mathbf{0.276}}{0.004}} \\
\midrule
\multicolumn{5}{@{}l}{\textbf{Qwen3-30B-A3B}} \\
$8$  & \figstatnum{\accerr{0.530}{0.015}} & \figstatnum{\accerr{0.538}{0.013}} & \figstatnum{\accerr{0.538}{0.014}} & \figstatnum{\accerr{\mathbf{0.595}}{0.016}} \\
$16$ & \figstatnum{\accerr{0.498}{0.010}} & \figstatnum{\accerr{0.521}{0.008}} & \figstatnum{\accerr{0.531}{0.012}} & \figstatnum{\accerr{\mathbf{0.563}}{0.021}} \\
$32$ & \figstatnum{\accerr{0.504}{0.005}} & \figstatnum{\accerr{0.493}{0.006}} & \figstatnum{\accerr{0.523}{0.008}} & \figstatnum{\accerr{\mathbf{0.588}}{0.007}} \\
\bottomrule
\end{tabular*}
\end{minipage}
\caption{Normalized training-score curves and summaries. Tables report the peak through step $256$ and the last-$128$ mean. Early peaks have no error bars; late-mean subscripts are temporal standard deviations. Bold marks each row's maximum; for late means, it also marks methods whose central value and the maximum each fall within the other's range.}
\label{fig:training_accuracy}
\begin{minipage}{\textwidth}
\centering
\includegraphics[width=0.84\textwidth]{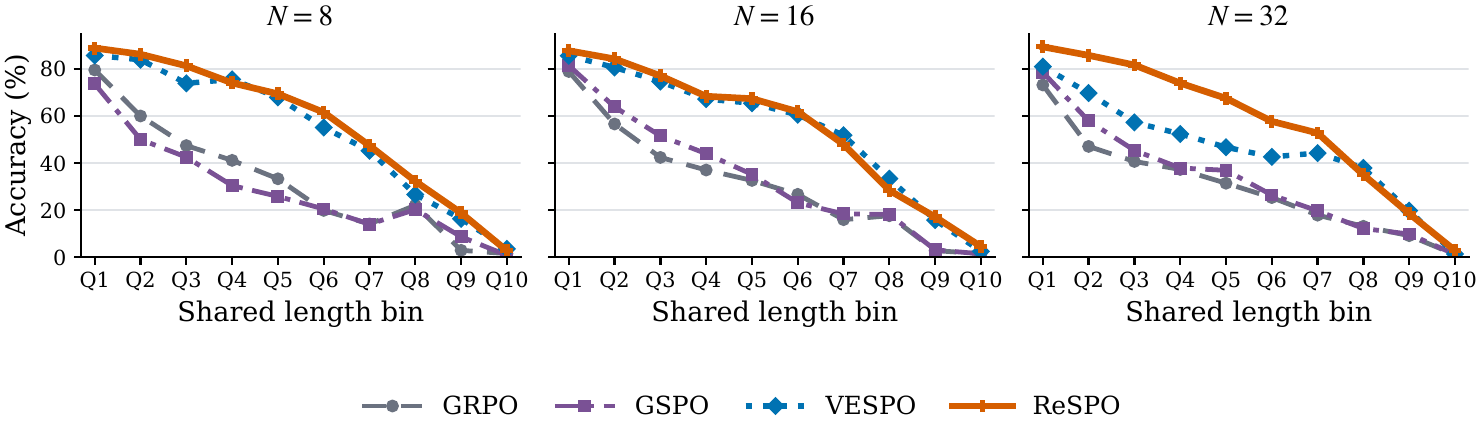}
\caption{1.7B accuracy across shared response-length bins. Q1--Q9 partition uncapped responses, while Q10 contains responses at the $16{,}384$-token cap. ReSPO has the highest accuracy in $8$ of $10$ bins at each $N$.}
\label{fig:length_quantiles_main}
\end{minipage}
\begin{minipage}{\textwidth}
\centering
\captionof{table}{Final-checkpoint pass@1. Bold marks the highest mean within each model--$N$ block.}
\label{tab:eval_results}
\fontsize{8}{8.4}\selectfont
\setlength{\tabcolsep}{3.2pt}
\renewcommand{\arraystretch}{0.92}
\begin{tabular}{@{}cllrrrrrr>{\columncolor{tableAverage}}r@{}}
\toprule
\rowcolor{tableHeader}
\textbf{Model} & \textbf{$N$} & \textbf{Method} & \textbf{AIME25} & \textbf{AIME24} & \textbf{AMC23} & \textbf{Olymp.} & \textbf{Minerva} & \textbf{MATH500} & \textbf{Avg.} \\
\midrule
\multirow{12}{*}{\shortstack{\textbf{Qwen3}\\\textbf{1.7B}}}
 & \multirow{4}{*}{8}  & \textsc{GRPO}  & 0.042 & 0.075 & 0.353 & 0.284 & 0.286 & 0.665 & $\accerr{0.284}{0.013}$ \\
 &                       & \textsc{GSPO}  & 0.062 & 0.077 & 0.402 & 0.304 & 0.272 & 0.658 & $\accerr{0.296}{0.013}$ \\
 &                       & \textsc{VESPO} & 0.073 & 0.127 & 0.478 & \textbf{0.388} & \textbf{0.307} & \textbf{0.742} & $\accerr{\mathbf{0.353}}{0.015}$ \\
 &                       & \respoCell{\textbf{\textsc{ReSPO}}} & \respoCell{\textbf{0.090}} & \respoCell{\textbf{0.133}} & \respoCell{\textbf{0.484}} & \respoCell{0.368} & \respoCell{0.289} & \respoCell{0.726} & \respoCell{$\accerr{0.348}{0.016}$} \\
\cmidrule(l){2-10}
 & \multirow{4}{*}{16} & \textsc{GRPO}  & 0.071 & 0.083 & 0.431 & 0.303 & 0.278 & 0.663 & $\accerr{0.305}{0.014}$ \\
 &                       & \textsc{GSPO}  & 0.046 & 0.075 & 0.419 & 0.319 & 0.283 & 0.672 & $\accerr{0.302}{0.013}$ \\
 &                       & \textsc{VESPO} & 0.065 & 0.106 & 0.464 & \textbf{0.364} & \textbf{0.301} & 0.712 & $\accerr{0.335}{0.014}$ \\
 &                       & \respoCell{\textbf{\textsc{ReSPO}}} & \respoCell{\textbf{0.094}} & \respoCell{\textbf{0.123}} & \respoCell{\textbf{0.467}} & \respoCell{0.358} & \respoCell{0.299} & \respoCell{\textbf{0.717}} & \respoCell{$\accerr{\mathbf{0.343}}{0.015}$} \\
\cmidrule(l){2-10}
 & \multirow{4}{*}{32} & \textsc{GRPO}  & 0.052 & 0.040 & 0.383 & 0.289 & 0.278 & 0.653 & $\accerr{0.283}{0.012}$ \\
 &                       & \textsc{GSPO}  & 0.046 & 0.054 & 0.381 & 0.299 & \textbf{0.299} & 0.663 & $\accerr{0.290}{0.013}$ \\
 &                       & \textsc{VESPO} & 0.081 & \textbf{0.131} & 0.456 & 0.341 & 0.282 & 0.698 & $\accerr{0.332}{0.015}$ \\
 &                       & \respoCell{\textbf{\textsc{ReSPO}}} & \respoCell{\textbf{0.088}} & \respoCell{0.117} & \respoCell{\textbf{0.488}} & \respoCell{\textbf{0.367}} & \respoCell{0.280} & \respoCell{\textbf{0.729}} & \respoCell{$\accerr{\mathbf{0.345}}{0.015}$} \\
\midrule
\multirow{12}{*}{\shortstack{\textbf{Qwen3}\\\textbf{30B-A3B}}}
 & \multirow{4}{*}{8}  & \textsc{GRPO}  & 0.152 & 0.248 & 0.738 & 0.511 & 0.436 & 0.852 & $\accerr{0.490}{0.018}$ \\
 &                       & \textsc{GSPO}  & 0.181 & 0.267 & 0.713 & \textbf{0.521} & 0.439 & \textbf{0.861} & $\accerr{\mathbf{0.497}}{0.018}$ \\
 &                       & \textsc{VESPO} & 0.154 & 0.273 & 0.700 & 0.517 & \textbf{0.454} & 0.853 & $\accerr{0.492}{0.018}$ \\
 &                       & \respoCell{\textbf{\textsc{ReSPO}}} & \respoCell{\textbf{0.252}} & \respoCell{\textbf{0.306}} & \respoCell{\textbf{0.791}} & \respoCell{0.427} & \respoCell{0.411} & \respoCell{0.783} & \respoCell{$\accerr{0.495}{0.019}$} \\
\cmidrule(l){2-10}
 & \multirow{4}{*}{16} & \textsc{GRPO}  & 0.138 & 0.246 & 0.733 & 0.490 & 0.424 & 0.831 & $\accerr{0.477}{0.017}$ \\
 &                       & \textsc{GSPO}  & 0.156 & 0.265 & 0.700 & 0.493 & 0.452 & 0.848 & $\accerr{0.486}{0.018}$ \\
 &                       & \textsc{VESPO} & 0.152 & 0.300 & 0.700 & \textbf{0.515} & 0.436 & 0.854 & $\accerr{0.493}{0.018}$ \\
 &                       & \respoCell{\textbf{\textsc{ReSPO}}} & \respoCell{\textbf{0.213}} & \respoCell{\textbf{0.321}} & \respoCell{\textbf{0.792}} & \respoCell{0.512} & \respoCell{\textbf{0.457}} & \respoCell{\textbf{0.857}} & \respoCell{$\accerr{\mathbf{0.525}}{0.018}$} \\
\cmidrule(l){2-10}
 & \multirow{4}{*}{32} & \textsc{GRPO}  & 0.123 & \textbf{0.313} & 0.717 & 0.503 & 0.444 & 0.850 & $\accerr{0.492}{0.017}$ \\
 &                       & \textsc{GSPO}  & 0.146 & 0.231 & 0.664 & 0.475 & 0.423 & 0.835 & $\accerr{0.462}{0.018}$ \\
 &                       & \textsc{VESPO} & 0.177 & 0.277 & 0.709 & 0.500 & 0.444 & \textbf{0.852} & $\accerr{0.493}{0.018}$ \\
 &                       & \respoCell{\textbf{\textsc{ReSPO}}} & \respoCell{\textbf{0.202}} & \respoCell{0.269} & \respoCell{\textbf{0.795}} & \respoCell{\textbf{0.510}} & \respoCell{\textbf{0.451}} & \respoCell{0.844} & \respoCell{$\accerr{\mathbf{0.512}}{0.018}$} \\
\bottomrule
\end{tabular}
\end{minipage}
\end{figure}
\endgroup

\textbf{Evaluation performance.}
At $N=8$, the macro averages are similar relative to their evaluation sampling uncertainty. For $N\in\{16,32\}$, ReSPO has the highest observed mean in all four model--$N$ settings, exceeding the highest baseline by $0.8$ and $1.3$ pp on 1.7B and by $3.2$ and $1.9$ pp on 30B. ReSPO has the highest value in $9$ of the $12$ model--benchmark comparisons at $N=16$ and $8$ at $N=32$. The pattern is most consistent on difficult competition benchmarks, where it has the highest AIME25 and AMC23 values in every model--$N$ setting. Averaged over AIME25, AIME24, and AMC23 on 30B, its differences from the highest baseline are $6.3$, $5.8$, and $3.4$ pp for $N\in\{8,16,32\}$, respectively. App.~\ref{app:detailed_comparisons} provides a detailed summary of these comparisons.

\textbf{Accuracy across response lengths.}
Using $10$ quantile-based response-length bins shared across methods and benchmarks for each $N$ (constructed in App.~\ref{app:length_accuracy}), accuracy decreases toward the long-response tail for all methods (Fig.~\ref{fig:length_quantiles_main}).
ReSPO has the highest pooled accuracy in $24$ of the $30$ bins. At $N=8$, it is highest in Q1--Q3 and Q5--Q9. At $N=16$, ReSPO is highest in Q1--Q6 and Q9--Q10. The lead is largest at $N=32$: ReSPO exceeds the highest baseline from Q1 through Q7 by $8.4$, $16.0$, $24.4$, $21.6$, $20.8$, $15.1$, and $8.5$ pp, respectively. The shared-bin comparison therefore shows that ReSPO's observed advantage persists within response-length ranges and is generally largest across short and medium responses at $N=32$.
In App.~\ref{app:length_accuracy}, Tab.~\ref{tab:length_quantile_metadata} gives the token boundaries and method-specific bin counts and Fig.~\ref{fig:length_quantiles} gives the complete per-benchmark results.

\subsection{Discussion}
\label{sec:discussion}

\textbf{Early learning from longer positive reasoning trajectories.}
The complete training trajectories in App.~\ref{app:training_stats} complement the positive-branch diagnostics in App.~\ref{app:w_diagnostics}. Particularly at $N=32$, ReSPO's response length and training score begin to separate from the baselines during the same early stage of optimization (Fig.~\ref{fig:stats_qwen17b_alpha1}). To examine this initial rapid improvement, App.~\ref{app:w_diagnostics} focuses on the first $512$ policy updates and the relationship between low-$W$ positive responses and rapid response-length growth. Because $\log W$ accumulates token-level policy drift across a response, longer responses tend to enter the low-$W$ tail. The diagnostic shows that ReSPO retains more coefficient mass in this tail while mean positive-response length grows faster. Together, these observations indicate that ReSPO can effectively learn from longer positive reasoning trajectories during the earliest training steps, when retaining their signal is most important for rapid improvement. On 30B, this early advantage remains $4.2$ and $5.4$ pp over the strongest baseline at $N=16$ and $N=32$, respectively, despite the soft length penalty. ReSPO also remains more accurate within shared response-length ranges (Fig.~\ref{fig:length_quantiles_main}), indicating that its advantage extends beyond the change in length distribution.

\textbf{Robustness across seeds.}
To assess the robustness of the observed differences at $N=32$, we augment each primary ReSPO run used in Fig.~\ref{fig:training_accuracy} and Tab.~\ref{tab:eval_results} with two additional seeds. All means and cross-seed population standard deviations below are computed over the three runs. On Qwen3-1.7B, the three runs attain a last-$128$ mean of $27.07\%$, with a standard deviation of $0.54$ pp. This mean exceeds VESPO ($23.00\%$), GRPO ($19.70\%$), and GSPO ($18.41\%$) by $4.07$, $7.37$, and $8.66$ pp.
On Qwen3-30B-A3B, the three runs attain a last-$128$ mean of $57.92\%$, with a standard deviation of $0.95$ pp. This mean exceeds VESPO ($52.30\%$), GRPO ($50.40\%$), and GSPO ($49.28\%$) by $5.62$, $7.52$, and $8.64$ pp. On the final-checkpoint evaluations averaged over AIME25, AIME24, and AMC23, the three-run mean is $41.94\%$ with a standard deviation of $0.47$ pp, exceeding VESPO ($38.77\%$), GRPO ($38.43\%$), and GSPO ($34.70\%$) by $3.17$, $3.51$, and $7.24$ pp. Across both model scales, the standard deviations are substantially smaller than ReSPO's performance gains over every baseline.

\textbf{Combination with Routing Replay.}
Routing Replay~\citep{ma2025routing}, which reuses rollout-time expert-routing decisions during optimization, is orthogonal to ReSPO's control of sequence-level importance weights. In a matched 30B model ablation at $N=16$, adding R3 increases the mean penalized training accuracy over the last $128$ policy steps from ${56.32\%}$ to ${58.16\%}$. In this comparison, Routing Replay improves late-stage training score and smoothness; App.~\ref{app:routing_replay} and Tabs.~\ref{tab:r3_training_metrics} report the complete comparison.
\section{Conclusion}

We identify a sign-dependent gradient-allocation problem in off-policy learning: clipping suppresses under-generated positive responses on the low-importance-weight tail while allowing severely over-generated negative responses to dominate on the high-weight tail. ReSPO addresses both failures with separate positive- and negative-advantage reshaping branches derived from an $\alpha$-divergence variational objective. The choice $\alpha^{(+)}>1$ preserves a nonzero positive-branch limit as $W\to0$, while the continuous $\alpha^{(-)}=1$ limit suppresses both negative-branch extremes.

Across Qwen3-1.7B and Qwen3-30B-A3B at $N\in\{8,16,32\}$, ReSPO attains the highest observed early score in five of six settings and the highest observed late score in five of six settings, with the observed differences generally largest at $N=32$. Positive-branch diagnostics show that ReSPO can effectively learn from long positive reasoning trajectories during the earliest training steps, even when accumulated policy drift might put them to the low-importance-weight tail; this early advantage remains influential on 30B despite its soft length penalty. ReSPO also obtains the highest observed final-checkpoint benchmark evaluation averages at $N=16$ and $N=32$ on both model scales, and has the highest accuracy in $24$ of $30$ response-length bins. The MoE ablation further shows that Routing Replay complements ReSPO by improving late-stage training score and smoothness. App.~\ref{app:limitations} discusses the limitations in compute budget and hyperparameter search.

\section*{Reproducibility Statement}
Sec.~\ref{sec:respo} defines the ReSPO objective, its gradient, and the analytical conditions used to select the kernel. App.~\ref{app:derivations} provides the full derivations, while App.~\ref{app:implementation} gives the numerically stable log-space implementation and complete algorithm. The accompanying code artifact contains the ReSPO implementation and launch configurations for the dense and MoE experiments. Tabs.~\ref{tab:hyperparams_training} and~\ref{tab:hyperparams_method} record the shared training configuration and every method-specific hyperparameter, including model backends, parallelism, optimization settings, sampling parameters, response limits, batch construction, rollout-reuse ratios, and training horizon.

Sec.~\ref{sec:experimental_setup} specifies the training data, model checkpoints, baseline configurations, and evaluation protocol. App.~\ref{app:dapo_reward} documents the DAPO-MATH reward and length penalty, and App.~\ref{app:validation_verifier} describes the evaluation-generation pipeline and mathematical-equivalence verifier. Tab.~\ref{tab:eval_results} reports every per-benchmark result used in the main comparison. App.~\ref{app:bootstrap_uncertainty} defines the evaluation resampling procedure and its random seed; Apps.~\ref{app:detailed_comparisons}, \ref{app:w_diagnostics}, \ref{app:on_policy_limit}, \ref{app:length_accuracy}, \ref{app:routing_replay}, and~\ref{app:training_stats} provide the detailed comparisons, sequence-weight diagnostics, on-policy limit, response-length analysis, Routing-Replay comparison, and auxiliary training statistics. Together, these materials specify how to reproduce the optimization method, training runs, evaluation scores, uncertainty estimates, and reported figures.

\bibliographystyle{iclr2027_conference}
\bibliography{references}

@article{neal2001ais,
  title={Annealed importance sampling},
  author={Neal, Radford M},
  journal={Statistics and Computing},
  volume={11},
  number={2},
  pages={125--139},
  year={2001},
  publisher={Springer}
}

@misc{zheng2025gspo,
      title={Group Sequence Policy Optimization}, 
      author={Chujie Zheng and Shixuan Liu and Mingze Li and Xiong-Hui Chen and Bowen Yu and Chang Gao and Kai Dang and Yuqiong Liu and Rui Men and An Yang and Jingren Zhou and Junyang Lin},
      year={2025},
      eprint={2507.18071},
      archivePrefix={arXiv},
      primaryClass={cs.LG},
      url={https://arxiv.org/abs/2507.18071}, 
}

@misc{shen2025vespo,
      title={{VESPO}: Variational Sequence-Level Soft Policy Optimization for Stable Off-Policy {LLM} Training},
      author={Guobin Shen and Chenxiao Zhao and Xiang Cheng and Lei Huang and Xing Yu},
      year={2026},
      eprint={2602.10693},
      archivePrefix={arXiv},
      primaryClass={cs.LG},
      url={https://arxiv.org/abs/2602.10693}, 
}

@inproceedings{lightman2024let,
  title={Let's Verify Step by Step},
  author={Lightman, Hunter and Kosaraju, Vineet and Burda, Yura and Edwards, Harri and Baker, Bowen and Lee, Teddy and Leike, Jan and Schulman, John and Sutskever, Ilya and Cobbe, Karl},
  booktitle={International Conference on Learning Representations},
  year={2024}
}

@article{chen2025learning,
  title={{ReSearch}: Learning to Reason with Search for {LLMs} via Reinforcement Learning},
  author={Chen, Mingyang and Sun, Linzhuang and Li, Tianpeng and Sun, Haoze and Zhou, Yijie and Zhu, Chenzheng and Wang, Haofen and Pan, Jeff Z and Zhang, Wen and Chen, Huajun and others},
  journal={arXiv preprint arXiv:2503.19470},
  year={2025}
}

@article{deepseek2025r1,
  title={{DeepSeek-R1}: Incentivizing Reasoning Capability in {LLMs} via Reinforcement Learning},
  author={{DeepSeek-AI} and Guo, Daya and Yang, Dejian and Zhang, Haowei and Song, Junxiao and Wang, Peiyi and Zhu, Qihao and Xu, Runxin and Zhang, Ruoyu and Ma, Shirong and Bi, Xiao and others},
  journal={arXiv preprint arXiv:2501.12948},
  year={2025}
}

@misc{qwen2025qwen3,
      title={{Qwen3} Technical Report},
      author={An Yang and Anfeng Li and Baosong Yang and Beichen Zhang and Binyuan Hui and Bo Zheng and Bowen Yu and Chang Gao and Chengen Huang and Chenxu Lv and Chujie Zheng and Dayiheng Liu and Fan Zhou and Fei Huang and Feng Hu and Hao Ge and Haoran Wei and Huan Lin and Jialong Tang and Jian Yang and Jianhong Tu and Jianwei Zhang and Jianxin Yang and Jiaxi Yang and Jing Zhou and Jingren Zhou and Junyang Lin and Kai Dang and Keqin Bao and Kexin Yang and Le Yu and Lianghao Deng and Mei Li and Mingfeng Xue and Mingze Li and Pei Zhang and Peng Wang and Qin Zhu and Rui Men and Ruize Gao and Shixuan Liu and Shuang Luo and Tianhao Li and Tianyi Tang and Wenbiao Yin and Xingzhang Ren and Xinyu Wang and Xinyu Zhang and Xuancheng Ren and Yang Fan and Yang Su and Yichang Zhang and Yinger Zhang and Yu Wan and Yuqiong Liu and Zekun Wang and Zeyu Cui and Zhenru Zhang and Zhipeng Zhou and Zihan Qiu},
      year={2025},
      eprint={2505.09388},
      archivePrefix={arXiv},
      primaryClass={cs.CL},
      url={https://arxiv.org/abs/2505.09388}, 
}

@article{shao2024deepseekmath,
  title={{DeepSeekMath}: Pushing the Limits of Mathematical Reasoning in Open Language Models},
  author={Shao, Zhihong and Wang, Peiyi and Zhu, Qihao and Xu, Runxin and Song, Junxiao and Bi, Xiao and Zhang, Haowei and Zhang, Mingchuan and Li, YK and Wu, Yang and others},
  journal={arXiv preprint arXiv:2402.03300},
  year={2024}
}

@article{schulman2017ppo,
  title={Proximal Policy Optimization Algorithms},
  author={Schulman, John and Wolski, Filip and Dhariwal, Prafulla and Radford, Alec and Klimov, Oleg},
  journal={arXiv preprint arXiv:1707.06347},
  year={2017}
}

@inproceedings{schulman2015trpo,
  title={Trust region policy optimization},
  author={Schulman, John and Levine, Sergey and Abbeel, Pieter and Jordan, Michael and Moritz, Philipp},
  booktitle={International Conference on Machine Learning},
  pages={1889--1897},
  year={2015},
  organization={PMLR}
}

@misc{yu2025dapo,
      title={{DAPO}: An Open-Source {LLM} Reinforcement Learning System at Scale},
      author={Qiying Yu and Zheng Zhang and Ruofei Zhu and Yufeng Yuan and Xiaochen Zuo and Yu Yue and Weinan Dai and Tiantian Fan and Gaohong Liu and Lingjun Liu and Xin Liu and Haibin Lin and Zhiqi Lin and Bole Ma and Guangming Sheng and Yuxuan Tong and Chi Zhang and Mofan Zhang and Wang Zhang and Hang Zhu and Jinhua Zhu and Jiaze Chen and Jiangjie Chen and Chengyi Wang and Hongli Yu and Yuxuan Song and Xiangpeng Wei and Hao Zhou and Jingjing Liu and Wei-Ying Ma and Ya-Qin Zhang and Lin Yan and Mu Qiao and Yonghui Wu and Mingxuan Wang},
      year={2025},
      eprint={2503.14476},
      archivePrefix={arXiv},
      primaryClass={cs.LG},
      url={https://arxiv.org/abs/2503.14476}, 
}

@article{hu2025reinforce,
  title={{REINFORCE++}: Stabilizing Critic-Free Policy Optimization with Global Advantage Normalization},
  author={Hu, Jian and Liu, Jason Klein and Xu, Haotian and Shen, Wei},
  journal={arXiv preprint arXiv:2501.03262},
  year={2025}
}

@inproceedings{noukhovitch2025async,
  title={Asynchronous {RLHF}: Faster and More Efficient Off-Policy {RL} for Language Models},
  author={Noukhovitch, Michael and Huang, Shengyi and Xhonneux, Sophie and Hosseini, Arian and Agarwal, Rishabh and Courville, Aaron},
  booktitle={International Conference on Learning Representations},
  year={2025},
  url={https://arxiv.org/abs/2410.18252}
}

@article{ma2025routing,
  title={Stabilizing {MoE} Reinforcement Learning by Aligning Training and Inference Routers},
  author={Ma, Wenhan and Zhang, Hailin and Zhao, Liang and Song, Yifan and Wang, Yudong and Sui, Zhifang and Luo, Fuli},
  journal={arXiv preprint arXiv:2510.11370},
  year={2025}
}

@misc{liu2025speed,
  title = {When Speed Kills Stability: Demystifying {RL} Collapse from the Training-Inference Mismatch},
  author = {Liu, Jiacai and Li, Yingru and Fu, Yuqian and Wang, Jiawei and Liu, Qian and Jiang, Zhuo},
  year = {2025},
  month = sep,
  url = {https://richardli.xyz/rl-collapse},
  note = {Web article, accessed September 25, 2026}
}

@misc{dwyer2025soft,
      title={It's Not You, It's Clipping: A Soft Trust-Region via Probability Smoothing for {LLM} {RL}},
      author={Madeleine Dwyer and Adam Sobey and Adriane Chapman},
      year={2025},
      eprint={2509.21282},
      archivePrefix={arXiv},
      primaryClass={cs.LG},
      url={https://arxiv.org/abs/2509.21282}, 
}

@misc{gao2025sapo,
      title={Soft Adaptive Policy Optimization}, 
      author={Chang Gao and Chujie Zheng and Xiong-Hui Chen and Kai Dang and Shixuan Liu and Bowen Yu and An Yang and Shuai Bai and Jingren Zhou and Junyang Lin},
      year={2025},
      eprint={2511.20347},
      archivePrefix={arXiv},
      primaryClass={cs.LG},
      url={https://arxiv.org/abs/2511.20347}, 
}

@article{metelli2018policy,
  title={Policy optimization via importance sampling},
  author={Metelli, Alberto Maria and Papini, Matteo and Faccio, Francesco and Restelli, Marcello},
  journal={Advances in Neural Information Processing Systems},
  volume={31},
  year={2018}
}

@article{ouyang2022training,
  title={Training language models to follow instructions with human feedback},
  author={Ouyang, Long and Wu, Jeffrey and Jiang, Xu and Almeida, Diogo and Wainwright, Carroll and Mishkin, Pamela and Zhang, Chong and Agarwal, Sandhini and Slama, Katarina and Ray, Alex and others},
  journal={Advances in Neural Information Processing Systems},
  volume={35},
  pages={27730--27744},
  year={2022}
}

@misc{zhai2025rewards,
      title={Rewards as Labels: Revisiting {RLVR} from a Classification Perspective},
      author={Zepeng Zhai and Meilin Chen and Jiaxuan Zhao and Junlang Qian and Lei Shen and Yuan Lu},
      year={2026},
      eprint={2602.05630},
      archivePrefix={arXiv},
      primaryClass={cs.LG},
      url={https://arxiv.org/abs/2602.05630}, 
}

@inproceedings{sheng2025verl,
  title={{HybridFlow}: A Flexible and Efficient {RLHF} Framework},
  author={Sheng, Guangming and Zhang, Chi and Ye, Zilingfeng and Wu, Xibin and Zhang, Wang and Zhang, Ru and Peng, Yanghua and Lin, Haibin and Wu, Chuan},
  booktitle={Proceedings of the Twentieth European Conference on Computer Systems},
  pages={1279--1297},
  year={2025}
}

@inproceedings{kwon2023pagedattention,
  title={Efficient Memory Management for Large Language Model Serving with {PagedAttention}},
  author={Kwon, Woosuk and Li, Zhuohan and Zhuang, Siyuan and Sheng, Ying and Zheng, Lianmin and Yu, Cody and Gonzalez, Joseph E. and Zhang, Hao and Stoica, Ion},
  booktitle={Proceedings of the 29th Symposium on Operating Systems Principles},
  pages={611--626},
  year={2023},
  doi={10.1145/3600006.3613165},
  url={https://doi.org/10.1145/3600006.3613165}
}

@misc{kydlicek2024mathverify,
  title={{Math-Verify}: Math Verification Library},
  author={Kydl{\'\i}{\v{c}}ek, Hynek},
  year={2024},
  howpublished={\url{https://github.com/huggingface/Math-Verify}},
  note={Software, accessed September 25, 2026}
}

@article{csiszar1975divergence,
  title={I-divergence geometry of probability distributions and minimization problems},
  author={Csisz{\'a}r, Imre},
  journal={The Annals of Probability},
  pages={146--158},
  year={1975},
  publisher={JSTOR}
}

@article{hendrycks2021math,
  title={Measuring Mathematical Problem Solving with the {MATH} Dataset},
  author={Hendrycks, Dan and Burns, Collin and Kadavath, Saurav and Arora, Akul and Basart, Steven and Tang, Eric and Song, Dawn and Steinhardt, Jacob},
  journal={arXiv preprint arXiv:2103.03874},
  year={2021}
}

@misc{aime2025,
  title={{AIME} 2025 Dataset Release},
  author={{OpenCompass}},
  year={2025},
  howpublished={\url{https://huggingface.co/datasets/opencompass/AIME2025}},
  note={Competition problems originally published by the Mathematical Association of America; accessed September 25, 2026}
}

@misc{aime2024,
  title={{AIME} 2024 Dataset Release},
  author={{Maxwell-Jia}},
  year={2024},
  howpublished={\url{https://huggingface.co/datasets/Maxwell-Jia/AIME_2024}},
  note={Competition problems originally published by the Mathematical Association of America; accessed September 25, 2026}
}

@misc{amc2023,
  title={{AMC} 2023 Dataset Release},
  author={{math-ai}},
  year={2023},
  howpublished={\url{https://huggingface.co/datasets/math-ai/amc23}},
  note={Competition problems originally published by the Mathematical Association of America; accessed September 25, 2026}
}

@inproceedings{he2024olympiadbench,
  title={{OlympiadBench}: A Challenging Benchmark for Promoting {AGI} with Olympiad-Level Bilingual Multimodal Scientific Problems},
  author={He, Chaoqun and Luo, Renjie and Bai, Yuzhuo and Hu, Shengding and Thai, Zhen and Shen, Junhao and Hu, Jinyi and Han, Xu and Huang, Yujie and Zhang, Yuxiang and others},
  booktitle={Proceedings of the 62nd Annual Meeting of the Association for Computational Linguistics (Volume 1: Long Papers)},
  pages={3828--3850},
  year={2024}
}

@article{lewkowycz2022minerva,
  title={Solving quantitative reasoning problems with language models},
  author={Lewkowycz, Aitor and Andreassen, Anders and Dohan, David and Dyer, Ethan and Michalewski, Henryk and Ramasesh, Vinay and Slone, Ambrose and Anil, Cem and Schlag, Imanol and Gutman-Solo, Theo and others},
  journal={Advances in Neural Information Processing Systems},
  volume={35},
  pages={3843--3857},
  year={2022}
}
\clearpage
\appendix
\raggedbottom
\section{Derivations for the \texorpdfstring{$\alpha$}{alpha} Reshaping Kernel}
\label{app:derivations}

This appendix collects the derivations referenced in Sec.~\ref{sec:respo}.

\subsection{\texorpdfstring{$\alpha$}{alpha}-Bregman divergence and first-order condition}
\label{app:renyi_fo}

For $f(x) = \frac{1}{\alpha(\alpha-1)} x^\alpha$ with $\alpha \neq 0, 1$, $f$ is strictly convex on $\mathbb{R}_+$. The induced Bregman divergence is
\begin{equation}\label{eq:renyi_div}
  D_f(p \,\|\, q) = \sum_{\bm{o}} \left[ \frac{1}{\alpha(\alpha-1)} p(\bm{o})^{\alpha} - \frac{1}{\alpha(\alpha-1)} q(\bm{o})^{\alpha} - \frac{1}{\alpha-1} q(\bm{o})^{\alpha-1} \big(p(\bm{o}) - q(\bm{o})\big) \right].
\end{equation}
The objective in Eq.~\eqref{eq:variational} has derivative
\[
  \frac{\partial}{\partial \tau(\bm{o})}\Big[ (1-\beta)\, D_f(\tau \,\|\, \mu) + \beta\, D_f(\tau \,\|\, \pi) \Big]
  = (1-\beta)\big(f'(\tau) - f'(\mu)\big) + \beta\big(f'(\tau) - f'(\pi)\big),
\]
so setting it to zero yields $f'(\tau^*) = (1-\beta)\,f'(\mu) + \beta\,f'(\pi)$. With $f'(x) = \frac{1}{\alpha-1} x^{\alpha-1}$ this reads
\begin{equation}
  \frac{1}{\alpha-1}(\tau^*)^{\alpha-1} = \frac{1-\beta}{\alpha-1}\, \mu^{\alpha-1} + \frac{\beta}{\alpha-1}\, \pi^{\alpha-1},
\end{equation}
and solving for $\tau^*$ gives the power-mean form of Eq.~\eqref{eq:tau_renyi}. Substituting $\pi = \mu\, W$ and using $\tau = \mu\,\phi(W)$ (Eq.~\eqref{eq:proposal}) recovers the unconstrained kernel Eq.~\eqref{eq:phi_unconstrained}. The normalized distribution is obtained only afterward as $\bar{\tau}=\tau/Z$.

\paragraph{What if we optimize over a normalized distribution?}
To distinguish this alternative from our unnormalized measure $\tau$, let $q$ denote the optimization variable. For $\alpha>1$, imposing $\sum_{\bm{o}}q(\bm{o}) = 1$ with multiplier $\eta$ gives the KKT conditions for $q \geq 0$: $$f'(q^*) = (1-\beta)\,f'(\mu) + \beta\,f'(\pi)-\eta,$$
which yield $$q^\star(\bm{o}) = [\max(0, (1-\beta)\mu(\bm{o})^{\alpha-1} + \beta\pi(\bm{o})^{\alpha-1} - \eta(\alpha-1))]^{1/(\alpha-1)},$$ including a positivity projection. Substituting $q = \mu\phi_0(W)$ and $\pi = \mu W$ gives
$$\phi_0\big(W; \mu(\bm{o})\big) = \left[ (1-\beta) + \beta W^{\alpha-1} - \eta(\alpha-1) \cdot \mu(\bm{o})^{1-\alpha}\right]^{1/(\alpha-1)}_+.$$
This loses the ratio-only form needed for a reshaping kernel. The normalization multiplier enters as $\eta(\alpha-1)\mu(\bm{o})^{1-\alpha}$ rather than as a constant, so the optimum depends on the pair $(W,\mu(\bm{o}))$ instead of on $W$ alone. Depending on the multiplier and local behavior mass, the positivity projection can also set the weight to zero, reintroducing a hard boundary. Optimizing over the unnormalized measure avoids both complications and yields the smooth kernel in Eq.~\eqref{eq:phi_unconstrained}.

\subsection{Positive branch: exponential tilt derivation}
\label{app:pos_lagrangian}

Given the unconstrained solution $\tau_0^*$ from the dual-proximity problem (Sec.~\ref{sec:variational}), we impose the second-moment constraint $\sum_{\bm{o}}\tau(\bm{o})\phi_0(W(\bm{o})) \leq C_1$ by solving the KL projection in Eq.~\eqref{eq:kl_proj}:
\begin{equation}
  \tau^* = \arg\min_{\tau}\; D_{\KL}(\tau \,\|\, \tau_0^*) \qquad \text{s.t.} \qquad \sum_{\bm{o}} \tau(\bm{o})\, \phi_0(W(\bm{o})) \leq C_1, \quad \sum_{\bm{o}} \tau(\bm{o})\leq C_2.
\end{equation}
Taking the functional derivative of the Lagrangian $\mathcal{L}(\tau, \lambda, \gamma) = D_{\KL}(\tau \| \tau_0^*) + \lambda(\sum_{\bm{o}}\tau(\bm{o})\phi_0(W(\bm{o})) - C_1) + \gamma (\sum_{\bm{o}}\tau(\bm{o})-C_2)$ gives
\begin{equation}
  \log \tau^*(\bm{o}) = \log \tau_0^*(\bm{o}) - \lambda\, \phi_0(W(\bm{o})) - \gamma ,
\end{equation}
and hence $\tau^*(\bm{o}) = \tau_0^*(\bm{o}) \cdot \exp(-\lambda\, \phi_0(W(\bm{o}))-\gamma)$, as stated in Eq.~\eqref{eq:exp_tilt}. This is the standard KL information projection under a linear constraint~\citep{csiszar1975divergence}: among all measures satisfying $\sum_{\bm{o}}\tau(\bm{o})\phi_0(W(\bm{o})) \leq C_1$, the one closest to $\tau_0^*$ in KL divergence is obtained by tilting the density with $\exp(-\lambda \phi_0(W))$. Substituting $\tau_0^* =\mu\,\phi_0(W)$ and extracting $\phi$ via $\tau^* = \mu\,\phi(W)$ gives
\begin{equation}
  \phi(W) = \phi_0(W) \exp(-\lambda\, \phi_0(W)-\gamma).
\end{equation}
The response-independent factor $\exp(-\gamma)$ changes only the total mass and can be absorbed into the constraint constants $C_1$ and $C_2$. Rescaling the kernel so that $\phi(1) = 1$, using $\phi_0(1) = 1$, gives Eq.~\eqref{eq:phi_exp}.

\paragraph{Why use the proxy $\phi_0$?}
The natural variance-control target is $\E_\mu[\phi(W)^2] \leq C$, which controls the second-moment term in the denominator of the standard importance-weight effective sample size. Since $\tau = \mu\,\phi$, this is equivalent to $\sum_{\bm{o}}\tau(\bm{o})\phi(W(\bm{o})) \leq C$. However, the final kernel $\phi$ is itself the unknown we are solving for: writing $\phi = \phi_0 \cdot h$ where $h$ is the tilt function, the constraint becomes $\sum_{\bm{o}}\tau(\bm{o})\phi_0(W(\bm{o}))h(W(\bm{o})) \leq C'$, which is \emph{not} linear in $\tau$ (since $h$ depends on $\tau$ through the measure-change identity). The Gibbs variational principle requires a constraint of the form $\sum_{\bm{o}}\tau(\bm{o})g(W(\bm{o})) \leq C$ for a \emph{fixed} function $g$.

We therefore replace $\phi$ with its known proxy $\phi_0$ (the unconstrained kernel from Step~1) and constrain $\sum_{\bm{o}}\tau(\bm{o})\phi_0(W(\bm{o})) \leq C$. This is linear in $\tau$, so the KL projection yields the clean exponential tilt $\exp(-\lambda \phi_0(W))$. Under this proxy, we prove that the derived $\tau$ satisfies bounded moments. Since the tilt satisfies $h(W) = \exp(\lambda(1 - \phi_0(W))) \leq \exp(\lambda)$, we have
\begin{equation}
  {\E_\mu[\phi(W)^2]} = \sum_{\bm{o}}\tau(\bm{o})\phi(W(\bm{o})) \leq \exp(\lambda)\sum_{\bm{o}}\tau(\bm{o})\phi_0(W(\bm{o})).
\end{equation}
Thus, bounding $\sum_{\bm{o}}\tau(\bm{o})\phi_0(W(\bm{o}))$ ensures $\E_\mu[\phi^2]$ is also bounded.

\paragraph{Why KL and not $\alpha$-divergence for the projection?}
One might ask why we use the KL divergence in the projection step (Step~2) rather than the same $\alpha$-divergence used in Step~1. Replacing the KL with $D_f(\tau \| \tau_0^*)$ for $f(x) = \frac{1}{\alpha(\alpha-1)}x^\alpha$ and using the linear constraint $\sum_{\bm{o}}\tau(\bm{o})\phi_0(W(\bm{o})) \leq C$, the first-order condition becomes
\begin{equation}
  \frac{1}{\alpha-1}\tau^{*\alpha-1} = \frac{1}{\alpha-1}\tau_0^{*\alpha-1} - \lambda\, \phi_0(W).
\end{equation}
Extracting $\phi$ yields the constraint term inside the power-mean bracket:
\begin{equation}
  \phi(W) \propto \Big[\phi_0(W)^{\alpha-1} - \lambda(\alpha-1)\, \mu^{1-\alpha}\, \phi_0(W)\Big]_+^{1/(\alpha-1)}.
\end{equation}
The factor $\mu^{1-\alpha}$ makes this expression depend on both $W$ and the behavior mass. For sufficiently small $\mu$, the bracket can become negative and activate the $[\cdot]_+$ projection; when $1<\alpha<2$, the subtracted term also grows faster than the first term as $\phi_0(W)$ increases. The resulting kernel can therefore acquire a hard cutoff rather than the smooth, ratio-only decay we seek.

The key distinction is geometric. The KL first-order condition is additive in \emph{log-space} ($\log\tau^* = \log\tau_0^* - \lambda\phi_0 -\gamma$), so the constraint enters as a \emph{multiplicative} exponential factor on $\tau_0^*$. In the power-Bregman alternative above, the constraint instead enters inside the power-mean bracket, introducing behavior-mass dependence and a possible hard boundary. The two-step procedure therefore uses each divergence where it is most natural: $\alpha$-divergence for reshaping (power-mean interpolation between $\mu$ and $\pi$), and KL for variance control (multiplicative exponential tilt).

\subsection{Hyperparameter derivations}
\label{app:hyperparam_derivations}

\paragraph{Positive branch: deriving $\lambda^{(+)}$.}
For $\phi(W) = \phi_0(W) \cdot \exp(\lambda(1 - \phi_0(W)))$, differentiating $\log\phi$ gives
\begin{equation}\label{eq:dlogphi_app}
  \frac{d \log \phi}{d W} = \phi_0'(W)\left(\frac{1}{\phi_0(W)} - \lambda\right).
\end{equation}
At $\alpha^{(+)} = 2$, the unconstrained kernel is $\phi_0^{(+)}(W) = (1-\beta^{(+)}) + \beta^{(+)} W$, so $\phi_0'(W) = \beta^{(+)} > 0$ is constant. At $W = 0$ we have $\phi_0(0) = 1 - \beta^{(+)}$, so
\[
  \frac{d\log\phi^{(+)}}{dW}\bigg|_{W=0} = \beta^{(+)}\!\left(\frac{1}{1-\beta^{(+)}} - \lambda^{(+)}\right).
\]
We additionally require the full kernel to be stationary at the origin, $\phi^{(+)}{}'(0)=0$. Because $\phi^{(+)}(0)>0$, this is equivalent to setting the log derivative above to zero, which gives
\begin{equation}\label{eq:eta_constraint}
  \lambda^{(+)} = \frac{1}{1-\beta^{(+)}}.
\end{equation}
For any $W > 0$, $\phi_0(W) > 1 - \beta^{(+)} = 1/\lambda^{(+)}$, so $1/\phi_0(W) < \lambda^{(+)}$ and $\frac{d\log\phi^{(+)}}{dW} < 0$: the kernel is strictly decreasing on its feasible domain $(0, \infty)$. Moreover, because $\alpha^{(+)}=2$ makes $\phi_0$ affine, the closed form has a differentiable extension to every $W\in\mathbb{R}$, even though importance weights themselves satisfy $W\geq0$. On this extension,
\[
  \phi^{(+)}{}'(W)
  =-\frac{(\beta^{(+)})^2}{1-\beta^{(+)}}\,W
  \exp\!\left(\lambda^{(+)}(1-\phi_0^{(+)}(W))\right),
\]
which is positive for $W<0$, zero at $W=0$, and negative for $W>0$. Thus $W=0$ is the unique global maximizer both on the real-valued extension and on the feasible domain, with
\begin{equation}
  \phi^{(+)}(0) = (1-\beta^{(+)})\, e^{\,\beta^{(+)}/(1-\beta^{(+)})}.
\end{equation}
With $\beta^{(+)} = 0.5$, this gives $\lambda^{(+)} = 2$ and $\phi^{(+)}(0) = \tfrac{1}{2}e^1 \approx 1.36$.

\paragraph{Why $\alpha=2$ is the natural choice.}
We seek a non-degenerate local condition at the origin: $\phi_0'(0)$ should be finite and strictly positive, so requiring $\phi^{(+)}{}'(0)=0$ uniquely determines $\lambda^{(+)}$. For $1<\alpha<2$, $\phi_0'(W)\propto W^{\alpha-2}$ is singular as $W\to0^+$; for $\alpha>2$, $\phi_0'(0)=0$, so stationarity of the full kernel holds for any $\lambda$ and does not determine the tilt. At $\alpha=2$, $\phi_0'(0)=\beta^{(+)}$ is finite and strictly positive, and $\phi_0$ is linear in $W$. Thus $\alpha^{(+)}=2$ uniquely combines an affine unconstrained kernel, a non-degenerate zero-derivative condition that fixes $\lambda^{(+)}$, and the real-line maximum at $W=0$ shown above.

\paragraph{Negative branch: matching the derivative at $W=1$.}
With $\alpha^{(-)}=1$ and $\beta^{(-)}=0.5$, the continuous-limit kernel is
\[
  \phi_0^{(-)}(W)=W^{\beta^{(-)}}=\sqrt{W}.
\]
We share $\beta=0.5$ across the two branches so that both unconstrained kernels interpolate symmetrically between the behavior and target policies. The remaining parameter $\lambda^{(-)}$ is fixed by matching the derivatives of the full kernels at the on-policy point. For the positive branch,
\[
  \phi^{(+)}(W)=\frac{1+W}{2}e^{1-W},
  \qquad
  \frac{d\phi^{(+)}}{dW}=-\frac{W}{2}e^{1-W},
\]
and hence $\phi^{(+)}{}'(1)=-\tfrac{1}{2}$. For the negative branch,
\[
  \phi^{(-)}(W)=W^{\beta^{(-)}}
  \exp\!\left(\lambda^{(-)}(1-W^{\beta^{(-)}})\right).
\]
Since $\phi^{(-)}(1)=1$, its derivative at $W=1$ is
\begin{equation}\label{eq:negative_slope}
  \phi^{(-)}{}'(1)=\beta^{(-)}\big(1-\lambda^{(-)}\big).
\end{equation}
Requiring $\phi^{(-)}{}'(1)=\phi^{(+)}{}'(1)$ and substituting $\beta^{(-)}=0.5$ gives
\[
  \frac{1}{2}\big(1-\lambda^{(-)}\big)=-\frac{1}{2}
  \qquad\Longrightarrow\qquad
  \lambda^{(-)}=2.
\]
Thus the positive and negative kernels have identical value and first derivative at $W=1$, making their local response to off-policy drift symmetric to first order.

\subsection{Continuous limit as \texorpdfstring{$\alpha \to 1$}{alpha to 1}}
\label{app:vespo_limit}

Taylor-expanding $W^{\alpha-1}$ around $\alpha = 1$ gives $W^{\alpha-1} = e^{(\alpha-1)\log W} \approx 1 + (\alpha-1)\log W$ to first order in $\alpha - 1$. Substituting this expansion into the $\alpha$ factor of Eq.~\eqref{eq:phi_exp} gives
\[
  \phi_0(W) = \big[ 1 + \beta(W^{\alpha-1} - 1) \big]^{1/(\alpha-1)} \approx \big[ 1 + \beta(\alpha-1)\log W \big]^{1/(\alpha-1)},
\]
which tends to $W^\beta$ as $\alpha \to 1$. The full kernel therefore reduces to $W^{\beta}\exp(\lambda(1-W^\beta))$, which is the form used by ReSPO's negative branch. For comparison, VESPO's gamma-IS kernel~\citep{shen2025vespo} is $W^{\beta}\exp(\lambda(1-W))$: the exponential tilt penalizes the raw importance weight $W$ rather than the reshaped weight $\phi_0(W)=W^\beta$.
\subsection{On-policy limit}
\label{app:on_policy_limit}

In the strict on-policy limit, $\pi_\theta=\pi_{\mathrm{old}}$ and every sample
has $W=1$. Setting $N=1$ removes staleness from rollout reuse and approximates
this condition. At $W=1$, the two kernels used in our
experiments satisfy
\begin{equation}
\phi^{(+)}(1)
= \frac{1+1}{2}\exp(1-1)=1,
\qquad
\phi^{(-)}(1)
= \sqrt{1}\exp\!\left(2(1-\sqrt{1})\right)=1.
\end{equation}
Consequently, ReSPO assigns unit reshaping weight in the exact on-policy
limit: its per-sequence multiplier is one. This recovers standard on-policy training, in which no importance-ratio weighting is applied.

\subsection{Length-dependent bias from sequence normalization}
\label{app:length_bias}

\citet{shen2025vespo} identify a length-dependent bias in the geometric-mean normalization used by GSPO. Let $T=|\bm{o}|$, let $\ell_t=\log w_t$ be the per-token log-ratio, and write $\bar{\ell}=T^{-1}\sum_{t=1}^{T}\ell_t$. The exact sequence importance weight and the GSPO weight are
\begin{equation}
  W=\exp(T\bar{\ell}),
  \qquad
  s=W^{1/T}=\exp(\bar{\ell}).
\end{equation}
The exponent $1/T$ controls variance, but it also changes the implied measure correction with response length. For a fixed length $T$, weighting samples from $\mu$ by $s$ induces the proposal
\begin{equation}
  Q_T(\bm{o})
  \propto \mu(\bm{o})s(\bm{o})
  = \mu(\bm{o})^{1-1/T}\pi(\bm{o})^{1/T}.
\end{equation}
Thus the target-policy exponent decreases as $T$ grows, and $Q_T$ is increasingly tempered toward the behavior policy $\mu$. In this sense, the strength of the off-policy correction depends on length rather than only on the likelihood shift between $\pi$ and $\mu$.

The normalization also conflates sequences with different total likelihood shifts. If two responses have lengths $T_1\neq T_2$ but the same average log-ratio $\bar{\ell}$, GSPO assigns both the same weight, $s_1=s_2=e^{\bar{\ell}}$. Their true sequence weights are instead $W_1=e^{T_1\bar{\ell}}$ and $W_2=e^{T_2\bar{\ell}}$, with
\begin{equation}
  \frac{W_2}{W_1}=\exp\!\big((T_2-T_1)\bar{\ell}\big).
\end{equation}
Consequently, a short response and a much longer response can receive the same GSPO weight even when their sequence-level distribution shifts differ exponentially. Clipping can further truncate these weights, but the length dependence is already introduced by the $1/T$ normalization.

ReSPO instead applies a fixed kernel $\phi(W)$ to the unnormalized sequence weight. Like any non-identity reshaping, this intentionally trades bias for variance. The same map $\phi$ is used at every response length, so length enters through the actual sequence likelihood ratio $W$ without an additional length-dependent tempering exponent.

\section{Log-Space Implementation of the Two-Branch Kernel}
\label{app:implementation}

The reshaping weight $\phi(W_i; \hat{A}_i)$ is evaluated per-sequence from the sequence-level log-ratio
\[
  \log W_i = \sum_{t=1}^{|\bm{o}_i|} \log w_{i,t}, \qquad w_{i,t} = \pitheta(o_{i,t} \mid \bm{q}, \bm{o}_{i,<t}) \,/\, \piold(o_{i,t} \mid \bm{q}, \bm{o}_{i,<t}).
\]
We clamp $\log W_i$ to $[-20, 20]$ before any non-linear operation and detach it from the gradient graph, since $\phi$ plays the role of a gradient scaling coefficient. The branch computations use numerically stable forms. The tail limits analyzed above refer to the unclamped analytical kernels; the implementation evaluates a finite-range approximation to avoid numerical overflow. We describe the two branches separately; which one applies to a given sequence is selected by the sign of $\hat{A}_i$.

\paragraph{Positive branch ($\hat{A}_i \geq 0$).}
With $\alpha^{(+)}=2$, $\beta^{(+)}=0.5$, $\lambda^{(+)}=2$, the kernel simplifies to
\begin{equation}\label{eq:log_phi_pos}
  \phi^{(+)}(W_i) = \frac{1+W_i}{2}\,e^{1-W_i}.
\end{equation}
No log-space tricks are needed: $(1+W_i)/2 \geq 1/2 > 0$ for all $W_i \geq 0$.

\paragraph{Negative branch ($\hat{A}_i < 0$).}
With $\alpha^{(-)}=1$, the continuous-limit kernel has $\phi_0^{(-)}(W_i)=W_i^{\beta^{(-)}}$. We therefore compute it directly in log-space:
\begin{equation}\label{eq:log_phi_neg}
  \log\phi_0^{(-)}(W_i)=\beta^{(-)}\log W_i.
\end{equation}
The full kernel adds the exponential tilt:
\begin{equation}
  \log\phi^{(-)}(W_i) = \log\phi_0^{(-)}(W_i) + \lambda^{(-)}\!\left(1 - \exp\!\left(\log\phi_0^{(-)}(W_i)\right)\right).
\end{equation}

\paragraph{Combining the branches.}
Given the two log-space scalars, we compute $\phi_i = \exp(\log\phi_i)$ on the branch selected by $\mathrm{sign}(\hat{A}_i)$, detach the result, and multiply it into the token-level loss as summarized in Alg.~\ref{alg:respo}. In practice we compute both branches for every sequence and select the applicable one with a sign mask, which is simpler on a GPU than using a data-dependent branch and has negligible overhead since both kernels are $O(1)$ per sequence.

\begin{algorithm}[t]
\caption{ReSPO: Reshaped Sequence Policy Optimization}
\label{alg:respo}
\begin{algorithmic}[1]
\REQUIRE Policy $\pitheta$, prompts $\{\bm{q}\}$, group size $G$
\REQUIRE $\alpha^{(+)}{=}2$, $\alpha^{(-)}{=}1$, $\beta^{(+)}{=}\beta^{(-)}{=}0.5$, $\lambda^{(+)}{=}2$, $\lambda^{(-)}{=}2$
\FOR{each training iteration}
  \STATE Sample a batch of prompts; for each $\bm{q}$, sample $G$ responses $\{\bm{o}_i\}_{i=1}^G \sim \piold$
  \STATE Compute rewards $R(\bm{q}, \bm{o}_i)$ and GRPO group-normalized advantages $\hat{A}_i$
  \FOR{each response $\bm{o}_i$}
    \STATE $\log W_i \leftarrow \sum_{t} \log\!\left[\pitheta(o_{i,t} \mid \bm{q}, \bm{o}_{i,<t}) / \piold(o_{i,t} \mid \bm{q}, \bm{o}_{i,<t})\right]$; clamp to $[-20, 20]$
    \IF{$\hat{A}_i \geq 0$}
      \STATE \COMMENT{positive branch}
      \STATE $\phi_i \leftarrow (1 + e^{\log W_i}) / 2 \cdot \exp(1 - e^{\log W_i})$
    \ELSE \STATE \COMMENT{negative branch}
      \STATE $\ell_i \leftarrow \beta^{(-)}\log W_i$
      \STATE $\phi_i \leftarrow \exp(\ell_i + \lambda^{(-)}(1 - e^{\ell_i}))$
    \ENDIF
    \STATE $\phi_i \leftarrow \phi_i$.\texttt{detach()}
  \ENDFOR
  \STATE $\nabla_\theta \mathcal{J} \leftarrow \frac{1}{G}\sum_i \phi_i \cdot \hat{A}_i \cdot \sum_t \nabla_\theta \log \pitheta(o_{i,t} \mid \bm{q}, \bm{o}_{i,<t})$
  \STATE Update $\theta$ by gradient ascent using $\nabla_\theta \mathcal{J}$
\ENDFOR
\end{algorithmic}
\end{algorithm}

\section{Detailed Hyperparameters}
\label{app:hyperparams}

Tab.~\ref{tab:hyperparams_training} lists the training hyperparameters for both experimental settings. Tab.~\ref{tab:hyperparams_method} lists the method-specific hyperparameters for each baseline and ReSPO.

\begin{table}[!ht]
\centering
\caption{Training hyperparameters across experimental settings.}
\label{tab:hyperparams_training}
\small
\begin{tabular}{lcc}
\toprule
Hyperparameter & \textbf{Qwen3-1.7B} & \textbf{Qwen3-30B-A3B} \\
\midrule
\multicolumn{3}{l}{\textit{Model \& infrastructure}} \\
Training backend & FSDP & Megatron \\
Tensor parallelism (train) & -- & 2 \\
Expert parallelism (train) & -- & 8 \\
Tensor parallelism (rollout) & 2 & 4 \\
\midrule
\multicolumn{3}{l}{\textit{Data}} \\
Training set & DAPO-MATH-17k & DAPO-MATH-17k \\
Max prompt length & $1{,}024$ & $1{,}024$ \\
Max response length & $15{,}360$ & $8{,}192$ \\
\midrule
\multicolumn{3}{l}{\textit{Optimization}} \\
Optimizer & AdamW & Adam \\
Learning rate & \multicolumn{2}{c}{$1 \times 10^{-6}$} \\
Warmup ratio & \multicolumn{2}{c}{$5\%$} \\
Weight decay & \multicolumn{2}{c}{$0.1$} \\
Gradient clip norm & \multicolumn{2}{c}{$1.0$} \\
KL penalty & \multicolumn{2}{c}{None} \\
\midrule
\multicolumn{3}{l}{\textit{Rollout}} \\
Rollout engine & \multicolumn{2}{c}{vLLM (async)} \\
Responses per prompt ($G$) & \multicolumn{2}{c}{$8$} \\
Temperature (train) & \multicolumn{2}{c}{$1.0$} \\
Top-$p$ (train) & \multicolumn{2}{c}{$1.0$} \\
\midrule
\multicolumn{3}{l}{\textit{Batching \& budget}} \\
Mini-batch size ($M$) & \multicolumn{2}{c}{$32$} \\
Rollout-reuse ratio ($N$) & \multicolumn{2}{c}{$\{8, 16, 32\}$} \\
Global batch size & \multicolumn{2}{c}{$N \times 32$} \\
Total policy updates & \multicolumn{2}{c}{$1024$} \\
Advantage estimator & \multicolumn{2}{c}{GRPO (group-normalized)} \\
Aggregation & \multicolumn{2}{c}{token-mean}\\
\bottomrule
\end{tabular}
\end{table}

\paragraph{Group-normalized advantages.}
The implementation computes GRPO advantages using the sample standard deviation with $\epsilon_A=10^{-6}$ added to the denominator. When all rewards in a group are identical, the centered numerators are zero, so all advantages are zero rather than undefined.

\paragraph{Why this normalization?}
For a fixed prompt, Eq.~\eqref{eq:measure_change} is an expectation over responses sampled from $\mu=\piold$. Given $G$ such responses, its standard Monte Carlo estimator is therefore the sequence mean $G^{-1}\sum_{i=1}^{G}$ used in Eq.~\eqref{eq:respo} and Alg.~\ref{alg:respo}; the outer expectation over prompts supplies the corresponding prompt average. The factor $1/G$ averages sampled sequences and does not normalize the unnormalized measure $\tau$ or its weights, so replacing it by $1/\sum_i\phi_i$ would instead produce a different, self-normalized estimator. In the reported experiments, we use the batch-wide token-mean normalization $1/\sum_i|\bm{o}_i|$ shown in Tab.~\ref{tab:hyperparams_training} for improved training stability. Relative to the sequence mean, this applies a common batch-level scale to all response contributions and therefore preserves their relative ReSPO weights within each batch.

\FloatBarrier
\begin{table}[!t]
\centering
\caption{Method-specific hyperparameters. All methods share the training configuration in Tab.~\ref{tab:hyperparams_training}.}
\label{tab:hyperparams_method}
\small
\begin{tabular}{llc}
\toprule
Method & Hyperparameter & Value \\
\midrule
GRPO & Clip ratio $\varepsilon$ & $0.2$ \\
\midrule
\multirow{2}{*}{GSPO} & Clip ratio low $\varepsilon_{\mathrm{low}}$ & $3 \times 10^{-4}$ \\
 & Clip ratio high $\varepsilon_{\mathrm{high}}$ & $4 \times 10^{-4}$ \\
\midrule
\multirow{4}{*}{VESPO}  & $\beta^{(+)}$ & $2$ \\
 & $\lambda^{(+)}$ & $3$ \\
 & $\beta^{(-)}$ & $3$ \\
 & $\lambda^{(-)}$ & $2$ \\
\midrule
\multirow{6}{*}{ReSPO} & $\alpha^{(+)}$ & $2$ \\
 & $\beta^{(+)}$ & $0.5$ \\
 & $\lambda^{(+)}$ & $2$ \\
 & $\alpha^{(-)}$ & $1$ \\
 & $\beta^{(-)}$ & $0.5$ \\
 & $\lambda^{(-)}$ & $2$ \\
\bottomrule
\end{tabular}
\end{table}

The VESPO values in Tab.~\ref{tab:hyperparams_method} are the best-performing
sign-specific configuration reported in the hyperparameter study of the
original VESPO paper~\citep{shen2025vespo}, and we use them directly. ReSPO
uses the analytically motivated parameters derived in
App.~\ref{app:derivations}.

\section{Evaluation Implementation}
In this section, we describe the training reward implementation and the evaluation scoring procedure.

\subsection{DAPO-MATH training reward implementation}
\label{app:dapo_reward}

In the reported runs, training examples carry the data-source key
\texttt{math\_dapo}, which selects the strict-box verifier rather than the
Math-Verify evaluator used for Tab.~\ref{tab:eval_results}. The relevant logic
is summarized below; names are simplified for clarity.
\begin{lstlisting}[style=pythoncode]
# Training-reward pseudocode
if use_strict_box_reward:
    strict_score = strict_box_verify(response, ground_truth)
    correct = (strict_score == +1)
    reward = +1.0 if correct else -1.0

if overlong_penalty_enabled:
    reward -= max(0.0, (response_length - 4096) / 4096)
\end{lstlisting}
The strict-box verifier returns $+1$ or $-1$ after comparing the ground truth
with the final boxed answer in the response suffix. Its wrapper explicitly
checks for $+1$ before mapping the result to the floating-point reward
$+1.0$ or $-1.0$. For the reported
30B experiments, the DAPO reward manager additionally applies the configured
overlong-buffer term.
With \texttt{max\_resp\_len}$=8{,}192$, buffer length $4{,}096$, and penalty factor
$1$, the additive penalty is
$-\max\{0,(L-4{,}096)/4{,}096\}$ for response length $L$. It is zero through $4{,}096$
tokens and reaches $-1$ at the response limit. The dense-model experiments
disable this term and use only the signed strict-box score.

\paragraph{Range of the penalized training accuracy.}
For the 1.7B experiments, the signed strict-box score is in $\{-1,1\}$, so
the affine transformation $(\texttt{score}+1)/2$ is in $\{0,1\}$. For the
30B experiments, the total score also includes the additive penalty in
$[-1,0]$, so the transformed quantity can range from $-0.5$ to $1$. We retain
the name \emph{penalized training accuracy} because the quantity reduces to
ordinary training accuracy when the penalty is zero and uses the same affine
scale; a negative value denotes an incorrect overlong response with an
additional length penalty, not a negative probability.

\subsection{Math-Verify scoring implementation}
\label{app:validation_verifier}

The evaluator tags all six benchmarks as \texttt{MATH-lighteval}, which the
\texttt{verl} score dispatcher routes to Math-Verify rather than the DAPO
strict-box verifier. The evaluation logic is summarized as follows.
\begin{lstlisting}[style=pythoncode]
# Evaluation-scoring pseudocode
gold = parse_latex(box(ground_truth))
predictions = parse(completion, modes=[expression, latex])
correct = any(
    mathematically_equivalent(g, p)
    for g in gold for p in predictions
)
score = 1.0 if correct else 0.0
\end{lstlisting}
Parsing the completion in both modes lets the verifier accept free-form final
answers in addition to strict \verb|\boxed{}| answers.
If either side cannot be parsed, no pair is mathematically equivalent, or the
scoring worker exceeds its $30$-second timeout, the completion receives zero
credit. The pass@1 reported in Tab.~\ref{tab:eval_results} is the mean of this
binary correctness indicator over samples and problems. The auxiliary
majority-vote metric separately uses the DAPO boxed-answer utilities.

\subsection{Bootstrap evaluation uncertainty}
\label{app:bootstrap_uncertainty}

The error bars in Tab.~\ref{tab:eval_results} use a cluster bootstrap over evaluation problems, using $B=10{,}000$ replicates with random seed $0$. For problem $i$ in benchmark $j$, let $y_{jik}\in\{0,1\}$ denote the correctness of sampled completion $k$. We first average the $n_{ji}$ completions for that problem,
\begin{equation}
z_{ji}=\frac{1}{n_{ji}}\sum_{k=1}^{n_{ji}}y_{jik},
\qquad
\widehat{a}_j=\frac{1}{m_j}\sum_{i=1}^{m_j}z_{ji},
\qquad
\widehat{A}=\frac{1}{6}\sum_{j=1}^{6}\widehat{a}_j,
\end{equation}
where $m_j$ is the number of problems in benchmark $j$: $30$ each for AIME25 and AIME24, $40$ for AMC23, $674$ for OlympiadBench, $272$ for MinervaMath, and $500$ for MATH-500. The reported macro average assigns equal weight to each benchmark.

For bootstrap replicate $b$, we independently draw $m_j$ problem indices with replacement within each benchmark, keeping all completions for a sampled problem together. If these indices are $I^{(b)}_{j1},\ldots,I^{(b)}_{jm_j}$, we recompute
\begin{equation}
\widehat{a}^{(b)}_j=\frac{1}{m_j}\sum_{\ell=1}^{m_j}z_{jI^{(b)}_{j\ell}},
\qquad
\widehat{A}^{(b)}=\frac{1}{6}\sum_{j=1}^{6}\widehat{a}^{(b)}_j.
\end{equation}
The subscripted error bar is the bootstrap standard error,
\begin{equation}
\mathrm{SE}_{\mathrm{boot}}
=\sqrt{\frac{1}{B-1}\sum_{b=1}^{B}\left(\widehat{A}^{(b)}-\overline{\widehat{A}}\right)^2},
\end{equation}
so an entry $\widehat{A}_{\pm\mathrm{SE}_{\mathrm{boot}}}$ reports one bootstrap standard error. For method contrasts, the same resampled problem indices are applied to both methods before taking the difference, giving a paired cluster bootstrap; percentile quantiles of that difference distribution form a confidence interval. These calculations quantify finite-evaluation sampling uncertainty conditional on the fixed final checkpoint and its stored generations.

\section{Additional Robustness Checks}
\label{app:additional_checks}

\subsection{Detailed training and evaluation comparisons}
\label{app:detailed_comparisons}

Within the first $256$ policy updates, ReSPO has the highest peak normalized
training score in five of six model--$N$ settings. The exception is 30B at
$N=8$, where GRPO is higher by $1.0$ pp. ReSPO's differences from the next-highest method
are
$2.3$, $1.0$, and $3.0$ pp on 1.7B for $N\in\{8,16,32\}$, respectively, and $4.2$
and $5.4$ pp on 30B at
$N=16,32$. Its curves remain above the baselines through most of the
subsequent trajectory.

Over the last $128$ policy steps, ReSPO has the highest mean in five of six
settings; on 1.7B at $N=16$, it is $0.4$ pp below VESPO, and each
central value lies within the other's temporal standard-deviation range. It
exceeds the highest clipped baseline in all six settings by $4.2$--$8.4$ pp,
with non-overlapping temporal ranges. From $N=8$ to $N=32$, its mean decreases
by $1.0$ pp on 1.7B and $0.7$ pp on 30B,
compared with $2.5$ and $4.5$ pp for GSPO and $4.5$ and $1.5$
pp for VESPO. At $N=32$, ReSPO exceeds the highest clipped
baseline by $7.9$ and $8.4$ pp, and VESPO by $4.6$ and $6.5$
pp. Across the early and late summaries, ReSPO is highest in
$10$ of $12$ settings. It is $1.0$ pp below the early maximum on 30B at $N=8$
and $0.4$ pp below the late maximum on 1.7B at $N=16$, where the temporal
ranges overlap.

At $N=8$, the evaluation macro averages are similar relative to their
bootstrap uncertainty: ReSPO is $0.5$ pp below VESPO on 1.7B
and $0.2$ pp below GSPO on 30B. For $N\in\{16,32\}$, ReSPO has the
highest observed mean in all four model--$N$ settings, exceeding the highest
baseline by $0.8$ and $1.3$ pp on 1.7B and by $3.2$ and $1.9$
pp on 30B. It has the highest value in $6$, $9$, and $8$ of the
$12$ model--benchmark pairs for $N\in\{8,16,32\}$, respectively.

The largest observed differences occur on difficult competition benchmarks.
On AIME25, ReSPO exceeds the highest baseline in all six model--$N$ settings
by $1.7$, $2.3$, and $0.7$ pp on 1.7B and by $7.1$, $5.7$,
and $2.5$ pp on 30B as $N$ increases. It also has the highest
AMC23 value in all six settings and the highest AIME24 value in four of six.
Averaged over these three benchmarks on 30B, its differences from the highest
baseline are $6.3$, $5.8$, and $3.4$ pp for $N\in\{8,16,32\}$, respectively. At
$N=8$, the other three benchmarks offset this difference in the macro average;
at larger $N$, ReSPO also attains the highest observed overall mean.

\subsection{Accuracy across response-length ranges}
\label{app:length_accuracy}

We analyze the stored Qwen3-1.7B evaluation generations underlying Tab.~\ref{tab:eval_results} for all four methods, all six benchmarks, and $N\in\{8,16,32\}$. Each method contributes $7{,}384$ responses per value of $N$, and response length is the number of Qwen3-1.7B tokenizer tokens in the generated text. For each value of $N$, we pool the observed lengths to construct one set of boundaries shared by every method and benchmark. The generation cap produces a point mass at $16{,}384$ tokens, so Q10 contains exactly the capped responses; Q1--Q9 divide the uncapped pooled responses into nine equal-frequency intervals. This tie-preserving construction keeps identical capped lengths in the same bin. Accuracy within a bin is the fraction of its responses accepted by the evaluation verifier.

Tab.~\ref{tab:length_quantile_metadata} reports the resulting token upper bounds and the pooled number of responses from each method in every bin. These counts expose the method-dependent response-length distributions used to compute the pooled accuracies below.


\begin{table}[!htbp]
\centering
\caption{Shared response-length-bin upper bounds (tokens) and pooled response counts for Qwen3-1.7B. Counts aggregate the six evaluation benchmarks; each method contributes $7{,}384$ responses for each $N$. Q10 contains responses at the $16{,}384$-token cap.}
\label{tab:length_quantile_metadata}
\setlength{\tabcolsep}{2.4pt}
\renewcommand{\arraystretch}{0.86}
\begin{tabular}{@{}clrrrrrrrrrr@{}}
\toprule
\rowcolor{tableHeader}
$N$ & Entry & Q1 & Q2 & Q3 & Q4 & Q5 & Q6 & Q7 & Q8 & Q9 & Q10 \\
\midrule
\rowcolor{tableAverage}
\multirow{5}{*}{$8$} & Upper bound & 355 & 480 & 611 & 761 & 954 & 1{,}263 & 2{,}009 & 4{,}404 & 16{,}383 & 16{,}384 \\
 & GRPO count & 1{,}043 & 908 & 820 & 855 & 816 & 809 & 573 & 163 & 102 & 1{,}295 \\
 & GSPO count & 1{,}253 & 1{,}087 & 1{,}131 & 1{,}074 & 917 & 775 & 420 & 79 & 46 & 602 \\
 & VESPO count & 386 & 466 & 456 & 475 & 539 & 671 & 1{,}037 & 1{,}504 & 1{,}376 & 474 \\
\respoMethodRow
 & \textbf{ReSPO} count & 207 & 429 & 482 & 482 & 605 & 630 & 856 & 1{,}140 & 1{,}360 & 1{,}193 \\
\midrule
\rowcolor{tableAverage}
\multirow{5}{*}{$16$} & Upper bound & 366 & 481 & 604 & 736 & 907 & 1{,}168 & 1{,}785 & 4{,}252 & 16{,}383 & 16{,}384 \\
 & GRPO count & 997 & 989 & 1{,}009 & 1{,}009 & 916 & 797 & 601 & 136 & 71 & 859 \\
 & GSPO count & 873 & 869 & 857 & 952 & 883 & 873 & 734 & 247 & 64 & 1{,}032 \\
 & VESPO count & 438 & 470 & 481 & 445 & 527 & 618 & 822 & 1{,}139 & 841 & 1{,}603 \\
\respoMethodRow
 & \textbf{ReSPO} count & 506 & 483 & 450 & 405 & 472 & 506 & 640 & 1{,}281 & 1{,}827 & 814 \\
\midrule
\rowcolor{tableAverage}
\multirow{5}{*}{$32$} & Upper bound & 371 & 483 & 596 & 711 & 856 & 1{,}057 & 1{,}421 & 2{,}732 & 16{,}383 & 16{,}384 \\
 & GRPO count & 1{,}172 & 1{,}013 & 1{,}000 & 906 & 849 & 698 & 542 & 289 & 109 & 806 \\
 & GSPO count & 949 & 908 & 874 & 905 & 831 & 832 & 730 & 386 & 81 & 888 \\
 & VESPO count & 516 & 556 & 555 & 605 & 693 & 762 & 795 & 973 & 975 & 954 \\
\respoMethodRow
 & \textbf{ReSPO} count & 255 & 394 & 453 & 469 & 499 & 589 & 805 & 1{,}230 & 1{,}714 & 976 \\
\bottomrule
\end{tabular}
\end{table}

\begin{figure}[!p]
\centering
\textbf{(a) Pooled benchmarks}\par\vspace{1pt}
\includegraphics[width=0.94\textwidth]{figures/response_length_quantiles_qwen17b.pdf}\par
\vspace{3pt}
\textbf{(b) AIME 2025, AIME 2024, and AMC 2023}\par\vspace{1pt}
\includegraphics[width=0.94\textwidth]{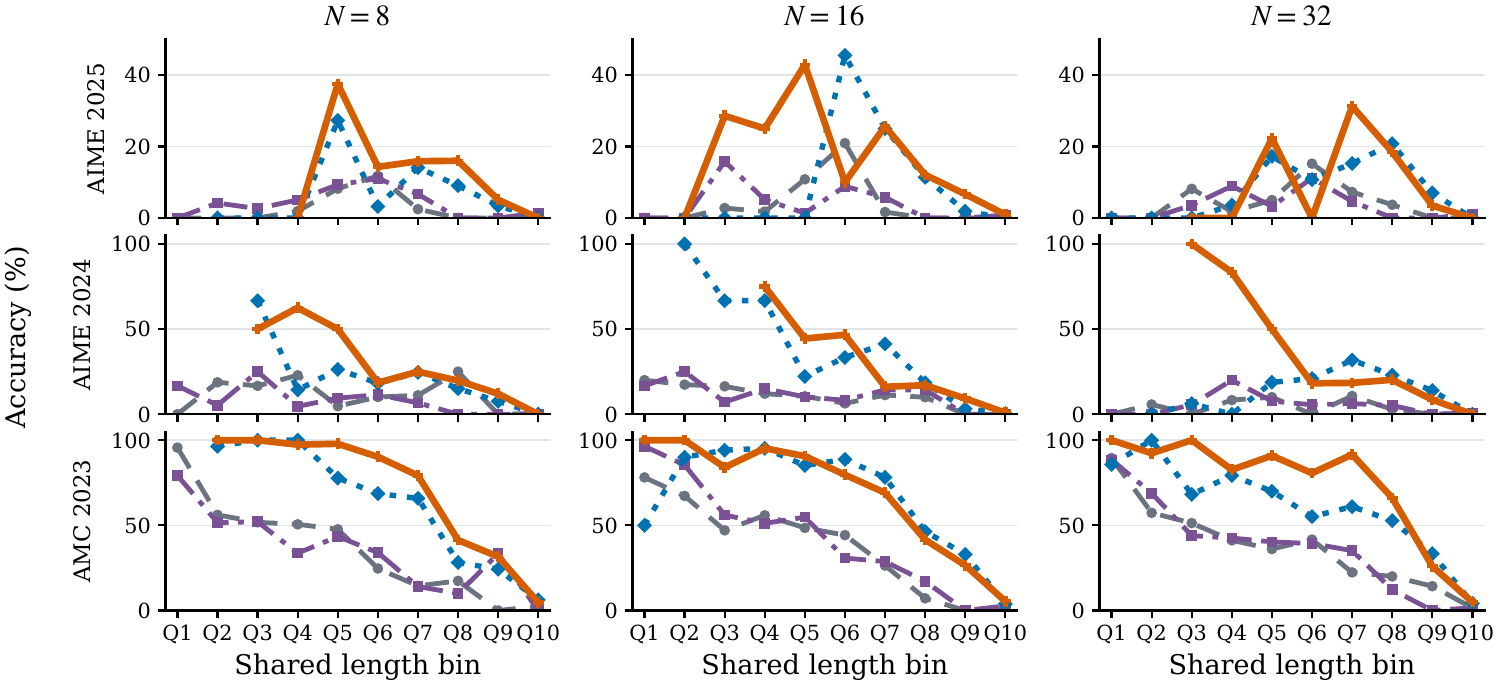}\par
\vspace{3pt}
\textbf{(c) OlympiadBench, MinervaMath, and MATH-500}\par\vspace{1pt}
\includegraphics[width=0.94\textwidth]{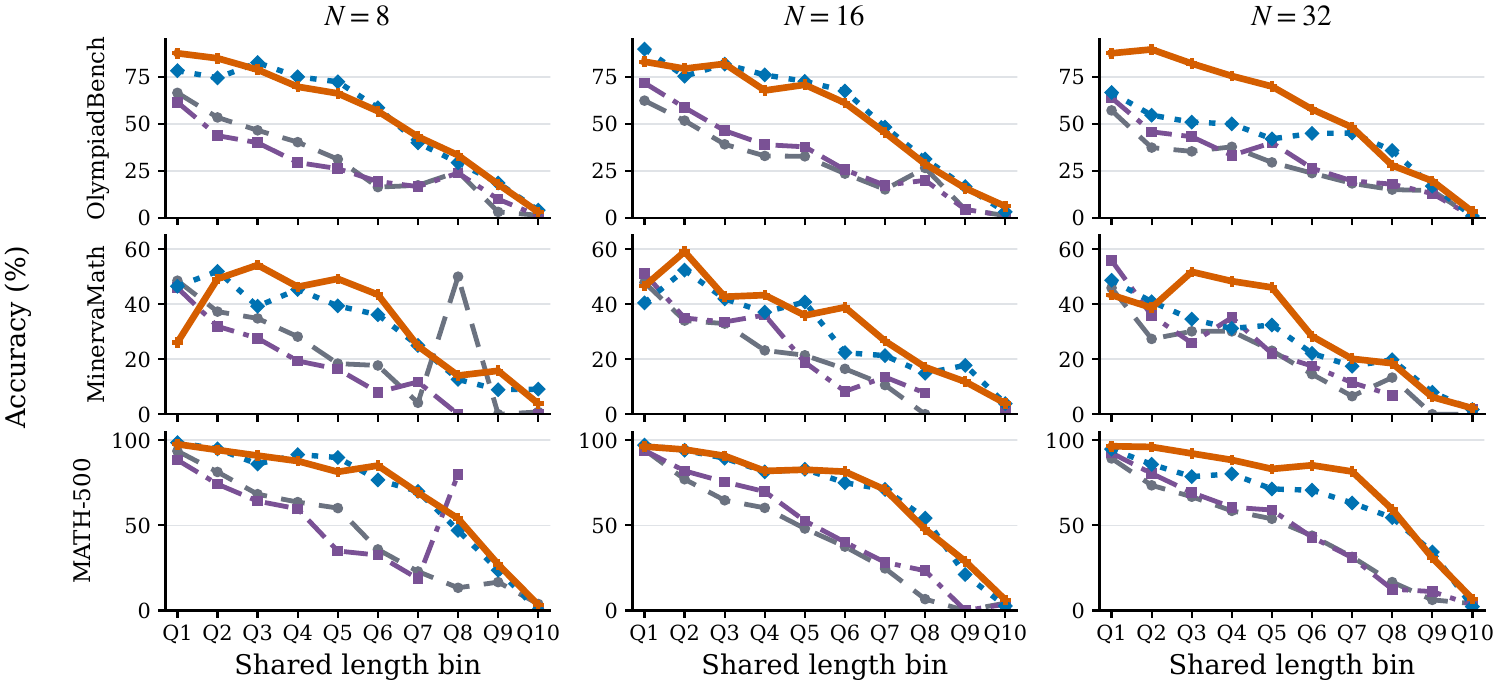}\par
\caption{Qwen3-1.7B accuracy by shared response-length bin. Q1--Q9 partition uncapped responses using boundaries shared across methods, while Q10 contains responses at the $16{,}384$-token cap. Columns correspond to $N\in\{8,16,32\}$; gaps in the benchmark panels indicate empty method--benchmark bins.}
\label{fig:length_quantiles}
\end{figure}

Fig.~\ref{fig:length_quantiles}(a) shows the pooled curves. Accuracy generally decreases toward the long tail, and the ReSPO curve decreases monotonically for all three values of $N$. ReSPO has the highest accuracy in eight of ten bins at each $N$. Its separation is widest at $N=32$: the differences from the highest baseline are $8.4$, $16.0$, $24.4$, $21.6$, $20.8$, $15.1$, and $8.5$ pp from Q1 through Q7. VESPO is higher in Q8 and Q9 by $2.9$ and $1.2$ pp, while ReSPO is higher in the cap bin by $1.5$ pp. At $N=8$ and $N=16$, the two reshaped methods are closer, but ReSPO still has the highest accuracy in most short, medium, and uncapped long-response bins.

Fig.~\ref{fig:length_quantiles}(b,c) separates the six benchmarks. The middle-length advantage at $N=32$ appears broadly on AMC 2023, OlympiadBench, and MATH-500. The AIME panels have fewer observations and lower base accuracy, producing more irregular bin-level curves. Across benchmarks, capped responses are generally among the least accurate groups.

\subsection{Positive-response weight and length dynamics}
\label{app:w_diagnostics}

Sec.~\ref{sec:discussion} suggests that ReSPO learns earlier from longer
positive responses. To isolate this initial rapid-improvement phase, the
diagnostic runs cover the first $16$ rollout batches, or $512$ of the $1024$
policy updates used in the complete experiments. Within this window, we ask
two questions: when a positive response moves into the low-$W$ tail, how much
learning coefficient does it retain, and how does this allocation evolve with
response length? Since $W=\pi_\theta(\bm{o})/\pi_{\mathrm{old}}(\bm{o})$,
$W<1$ means that the response has become less likely under the updated policy.
For $\hat A>0$, this is precisely a response whose probability the update
should increase.

We measure $W$ directly in matched Qwen3-1.7B-Base runs at $N=32$ with a
$16{,}384$-token response limit. ReSPO and VESPO use the same DAPO-MATH data,
optimizer settings, GRPO advantages, and token-mean aggregation. Each run
uses this first-half diagnostic window. For every sequence, we record its
advantage sign and
$\log W=\sum_t(\log\pi_\theta-\log\pi_{\mathrm{old}})$ in fixed bins. Because
this sum accumulates policy drift token by token, longer responses provide more
opportunities for negative drift to build up and therefore tend to occur in
the low-$W$ tail. For a positive-branch bin $B$, we report
\[
p_B^{(+)}=
\frac{\sum_{i:\,\hat A_i>0}\mathbf{1}\{\log W_i\in B\}}
     {\sum_{i:\,\hat A_i>0}1},
\qquad
m_B^{(+)}=
\frac{\sum_{i:\,\hat A_i>0}|\hat A_i|\phi^{(+)}(W_i)
      \mathbf{1}\{\log W_i\in B\}}
     {\sum_{i:\,\hat A_i>0}|\hat A_i|\phi^{(+)}(W_i)}.
\]
Here, $p_B^{(+)}$ is the fraction of positive responses in the bin, whereas
$m_B^{(+)}$ is the fraction of the total positive-branch loss coefficient
$|\hat A|\phi^{(+)}(W)$ assigned to that bin. Their ratio
$m_B^{(+)}/p_B^{(+)}$ is the amplification: a value above $1$ means that each
response in the bin receives more coefficient mass than the positive-branch
average, while a value below $1$ indicates suppression. The diagnostic logs
each numerator and denominator under the same distributed reduction, so these
ratios remain directly comparable.

\begin{table}[!htbp]
\centering
\caption{Positive-branch deep-tail allocation on Qwen3-1.7B-Base at
$N=32$. Values are means over rollout batches $9$--$16$; amplification is the
coefficient-mass share divided by the sequence share.}
\label{tab:w_diagnostics}
\small
\begin{tabular}{lrr}
\toprule
Statistic & ReSPO & VESPO \\
\midrule
Positive responses with $\log W<-1$ & $22.8\%$ & $16.7\%$ \\
Coefficient mass on $\log W<-1$ & $38.5\%$ & $15.5\%$ \\
Amplification on $\log W<-1$ & $1.69\times$ & $0.93\times$ \\
\midrule
Positive responses in $-5<\log W\leq-2$ & $7.4\%$ & $3.7\%$ \\
Coefficient mass in $-5<\log W\leq-2$ & $14.8\%$ & $1.1\%$ \\
Amplification in $-5<\log W\leq-2$ & $2.00\times$ & $0.30\times$ \\
\bottomrule
\end{tabular}
\end{table}

The distinction between response share and coefficient-mass share isolates
where the methods differ. In $\log W<-1$, ReSPO observes $22.8\%$ of its
positive responses but assigns them $38.5\%$ of its positive-branch
coefficient mass, giving $1.69\times$ amplification. VESPO observes a similar
$16.7\%$ response share but assigns the region only $15.5\%$ of its mass. The
contrast sharpens in $-5<\log W\leq-2$: ReSPO amplifies the bin to
$2.00\times$ its response share, whereas VESPO suppresses it to $0.30\times$.
At the final rollout batch, ReSPO places $58.5\%$ of its positive-branch mass
in $-5<\log W\leq-0.5$; VESPO instead places its largest share, $35.6\%$, in
the region immediately below $W=1$, $-0.5<\log W\leq0$.

\begin{table}[!htbp]
\centering
\caption{Mean $\log W$ and response length on the positive branch during the
$N=32$ diagnostic runs.}
\label{tab:w_diagnostic_trajectory}
\small
\begin{tabular}{crrrr}
\toprule
& \multicolumn{2}{c}{Mean $\log W$}
& \multicolumn{2}{c}{Mean response length} \\
\cmidrule(lr){2-3}\cmidrule(lr){4-5}
Batch & ReSPO & VESPO & ReSPO & VESPO \\
\midrule
$1$  & $-0.021$ & $+0.134$ & $752$  & $829$ \\
$8$  & $-0.188$ & $-0.064$ & $829$  & $843$ \\
$16$ & $-0.385$ & $-0.064$ & $1076$ & $915$ \\
\bottomrule
\end{tabular}
\end{table}

The weight and length statistics evolve together. By batch $16$, mean
$\log W$ reaches $-0.385$ for ReSPO and $-0.064$ for VESPO, corresponding to
geometric-mean weights of approximately $0.68$ and $0.94$. Over the same
interval, mean positive-response length grows from $752$ to $1076$ under ReSPO
($43\%$) and from $829$ to $915$ under VESPO ($10\%$). Thus, the faster growth
of ReSPO's positive responses coincides with their movement farther into the
low-$W$ region where ReSPO retains more coefficient mass. In the complete
$1024$-update trajectories, ReSPO's aggregate response length and training
score also separate from the baselines during the early stage of optimization
and continue to diverge at $N=32$ (Fig.~\ref{fig:stats_qwen17b_alpha1}). The
two views together are consistent with ReSPO learning earlier from longer
positive responses: this diagnostic identifies the retained learning
coefficient, while App.~\ref{app:training_stats} shows the corresponding
full-training trajectory.

To test the length--weight relationship directly, we also record positive-response length within each $\log W$ bin over all
$16$ rollout batches.
Tab.~\ref{tab:w_length_by_bin} reports pooled, count-weighted mean lengths and
normalizes each value by the corresponding method's positive-branch mean over
the same window. For $\log W\leq-0.5$, positive responses average $1276$ tokens
under ReSPO and $936$ under VESPO, or $1.13\times$ and $1.06\times$ their
branch-wide means, respectively. These pooled regions contain $7163$ and
$4918$ positive responses. Responses immediately below $W=1$ are shorter, so
the relation turns upward in the deep low-$W$ tail rather than changing
monotonically. Individual extreme-bin means are descriptive because those bins
are sparse. Combined with Tab.~\ref{tab:w_diagnostics}, the pooled result shows
that the kernels differ in how much learning-coefficient mass they retain on
these longer positive responses.

\begin{table}[!htbp]
\centering
\caption{Count-weighted positive-response length by sequence-weight bin over
rollout batches $1$--$16$ of the matched Qwen3-1.7B-Base follow-up runs at
$N=32$ and a $16{,}384$-token response limit. Parentheses give length relative
to the corresponding method's positive-branch mean over the same window;
$n$ is the positive-response count and ``--'' denotes an empty bin.}
\label{tab:w_length_by_bin}
{\small
\setlength{\tabcolsep}{3.5pt}
\begin{tabular}{lcrrrr}
\toprule
& & \multicolumn{2}{c}{ReSPO} & \multicolumn{2}{c}{VESPO} \\
\cmidrule(lr){3-4}\cmidrule(lr){5-6}
$\log W$ bin & $W$ range & Length & $n$ & Length & $n$ \\
\midrule
$(-\infty,-20]$ & $(0,2.1{\times}10^{-9}]$ & $2155\;(1.90\times)$ & $9$ & -- & $0$ \\
$(-20,-10]$ & $2.1{\times}10^{-9}$--$4.5{\times}10^{-5}$ & $2208\;(1.95\times)$ & $52$ & -- & $0$ \\
$(-10,-5]$  & $4.5{\times}10^{-5}$--$0.007$ & $2227\;(1.97\times)$ & $240$ & $5593\;(6.35\times)$ & $8$ \\
$(-5,-2]$   & $0.007$--$0.14$ & $1501\;(1.32\times)$ & $1687$ & $1085\;(1.23\times)$ & $592$ \\
$(-2,-1]$   & $0.14$--$0.37$ & $1199\;(1.06\times)$ & $2753$ & $957\;(1.09\times)$ & $2071$ \\
$(-1,-0.5]$ & $0.37$--$0.61$ & $1090\;(0.96\times)$ & $2422$ & $861\;(0.98\times)$ & $2247$ \\
$(-0.5,0]$  & $0.61$--$1$ & $885\;(0.78\times)$ & $3867$ & $759\;(0.86\times)$ & $3862$ \\
$(0,0.5]$   & $1$--$1.65$ & $994\;(0.88\times)$ & $2877$ & $840\;(0.95\times)$ & $3090$ \\
$(0.5,1]$   & $1.65$--$2.7$ & $1042\;(0.92\times)$ & $2036$ & $878\;(1.00\times)$ & $2060$ \\
$(1,2]$     & $2.7$--$7.4$ & $1202\;(1.06\times)$ & $1794$ & $981\;(1.11\times)$ & $1717$ \\
$(2,5]$     & $7.4$--$148$ & $1835\;(1.62\times)$ & $535$ & $1259\;(1.43\times)$ & $364$ \\
$(5,10]$    & $148$--$2.2{\times}10^4$ & $4985\;(4.40\times)$ & $6$ & $13430\;(15.26\times)$ & $1$ \\
$(10,20]$   & $2.2{\times}10^4$--$4.9{\times}10^8$ & -- & $0$ & -- & $0$ \\
$(20,\infty)$ & $>4.9{\times}10^8$ & -- & $0$ & -- & $0$ \\
\midrule
Positive-branch mean & -- & $1133$ & $18{,}278$ & $880$ & $16{,}012$ \\
\bottomrule
\end{tabular}
}
\end{table}

The allocation follows directly from the positive-branch kernels. ReSPO uses
\[
\phi^{(+)}(W)=\frac{1+W}{2}\exp(1-W),
\]
which approaches $e/2\approx1.359$ as $W\to0$ and remains $1.348$ at
$\log W=-2$. VESPO uses
$\phi^{(+)}(W)=W^2\exp(3(1-W))$, which vanishes as $W\to0$ and falls to
$0.245$ at $\log W=-2$ and $0.006$ at $\log W=-4$. At the same low value of
$W$, ReSPO therefore retains a near-maximal coefficient on a positive response,
whereas VESPO progressively removes its contribution. This kernel difference
explains the coefficient-mass allocation in Tab.~\ref{tab:w_diagnostics}.

\subsection{MoE training and Routing Replay}
\label{app:routing_replay}

Both Qwen3-30B-A3B ReSPO runs at $N=16$ reach the full budget of $1024$
policy updates. The Routing-Replay ablation adds R3 while matching the model,
data, optimization, response cap, batch sizes, and update budget. Both use the
current ReSPO parameters
$(\alpha^{(+)},\alpha^{(-)})=(2,1)$,
$(\beta^{(+)},\beta^{(-)})=(0.5,0.5)$, and
$(\lambda^{(+)},\lambda^{(-)})=(2,2)$.

Tab.~\ref{tab:r3_training_metrics} uses the same reporting protocol as
Fig.~\ref{fig:training_accuracy}. ``Early peak'' is the maximum through
policy step $256$; late metrics are policy-step-weighted trapezoidal means over
$[896,1024]$; and ``final'' is the value at step $1024$. The score is
$(\texttt{critic/score/mean}+1)/2$. Because the 30B reward includes the soft
overlong penalty described in App.~\ref{app:dapo_reward}, this is a
penalized training accuracy. The
subscript on the late score denotes its within-run temporal standard deviation.

\begin{table}[!ht]
\centering
\caption{Matched Qwen3-30B-A3B ReSPO training diagnostics at $N=16$, with and without R3 Routing Replay. All columns except the early peak and final score are time-weighted means over policy steps $[896,1024]$.}
\label{tab:r3_training_metrics}
\small
\begin{tabular}{lrrr}
\toprule
Setting & Early peak & Late score & Final score \\
\midrule
ReSPO
& $0.445$ & $\accerr{0.563}{0.021}$ & $0.555$ \\
ReSPO $+$ R3
& $0.446$ & $\accerr{0.582}{0.008}$ & $0.584$ \\
\bottomrule
\end{tabular}

\vspace{4pt}
\begin{tabular}{lrrrrr}
\toprule
Setting & Resp. len. & Cap-hit rate & Approx. KL & Entropy & Grad. norm \\
\midrule
ReSPO
& $1992$ & $3.505\%$ & $3.37{\times}10^{-4}$ & $0.128$ & $1.13$ \\
ReSPO $+$ R3
& $1213$ & $0.011\%$ & $4.77{\times}10^{-4}$ & $0.114$ & $1.21$ \\
\bottomrule
\end{tabular}
\end{table}

R3 leaves the early peak essentially unchanged but raises the late score by
$1.84$ pp and reduces its temporal standard deviation from
$2.11$ to $0.75$ pp. It also substantially shortens responses
and nearly eliminates response-cap hits in this window. Approximate KL rises
slightly. In this matched $N=16$ run, R3 combines with ReSPO, raising the
late-stage score and reducing its temporal variation.

\section{Limitations}
\label{app:limitations}

\textbf{Compute resource limitation.}
Our compute budget does not support a longer Qwen3-30B-A3B study with a
longer training response limit and additional policy updates. The
reported 30B experiments instead use an $8{,}192$-token limit with the soft
length penalty described in Sec.~\ref{sec:experimental_setup}. Under our
current training setup, extending both the context length and training horizon
could require several weeks on $8\times$H200 GPUs. Owing to the same compute constraints, we run multiple ReSPO seeds only for
$N=32$, where the separation from the baselines is largest and robustness is
most important to verify.

\textbf{Hyperparameter optimality.}
The ReSPO hyperparameters are selected primarily based on the analytical tail
requirements and derivative matching at $W=1$. This study does not exhaustively search
over $\alpha$, $\beta$, and $\lambda$; other choices may further improve
particular model scales, reuse ratios, or response-length regimes. However, we want to emphasize that without extensive hyperparameter tuniing, matching the tail requirements only is enough to find a good algorithm.

\section{Detailed Training Statistics and Reproducibility}
\label{app:training_stats}

\begin{figure}[!ht]
\centering
\includegraphics[width=0.98\textwidth]{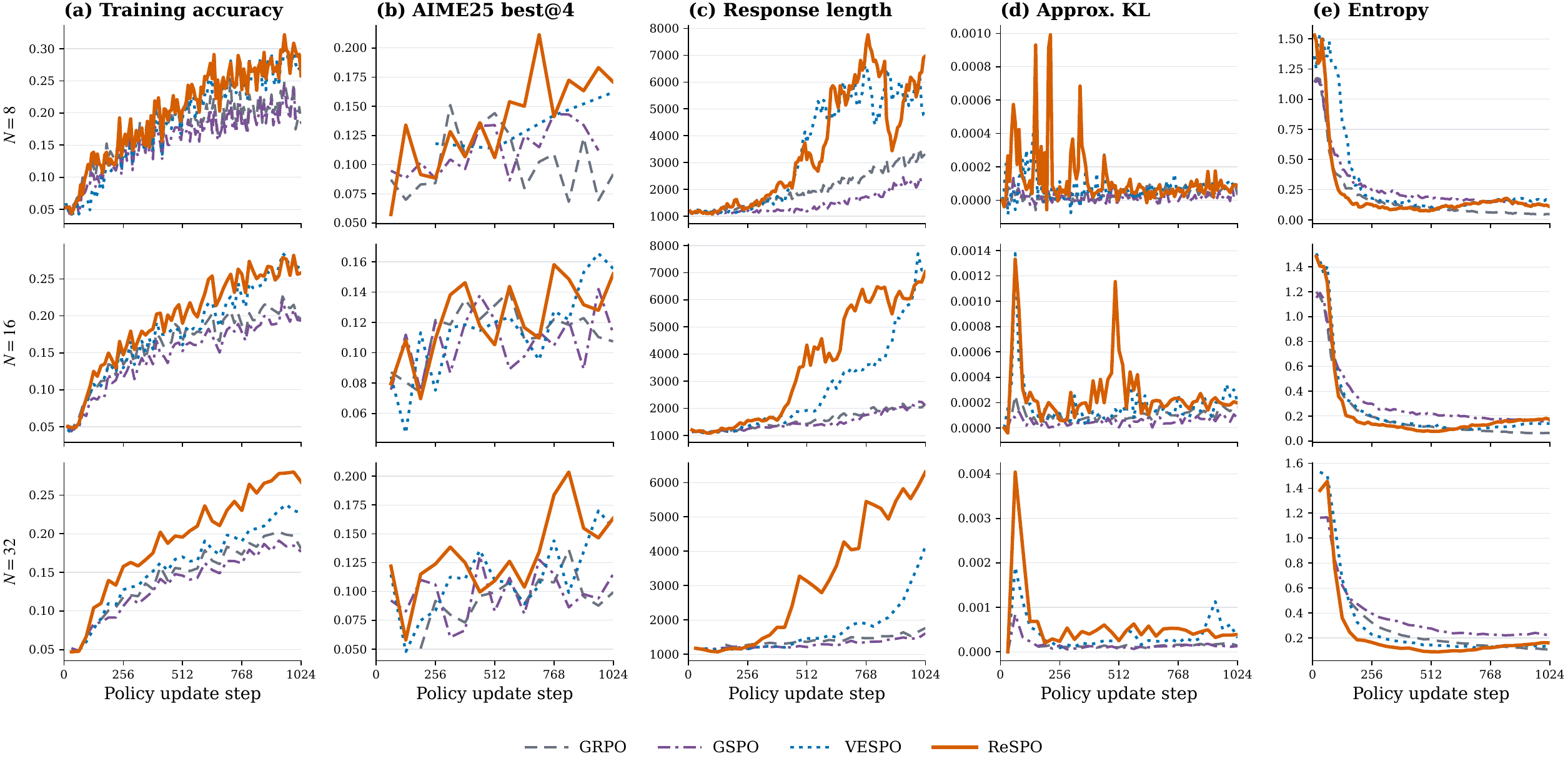}
\caption{Detailed training statistics for Qwen3-1.7B-Base on DAPO-MATH. Rows correspond to $N\in\{8,16,32\}$. Training accuracy is $(\texttt{critic/score/mean}+1)/2$, and the evaluation panel reports AIME25 best@4 accuracy.}
\label{fig:stats_qwen17b_alpha1}
\end{figure}

\begin{figure}[!ht]
\centering
\includegraphics[width=0.98\textwidth]{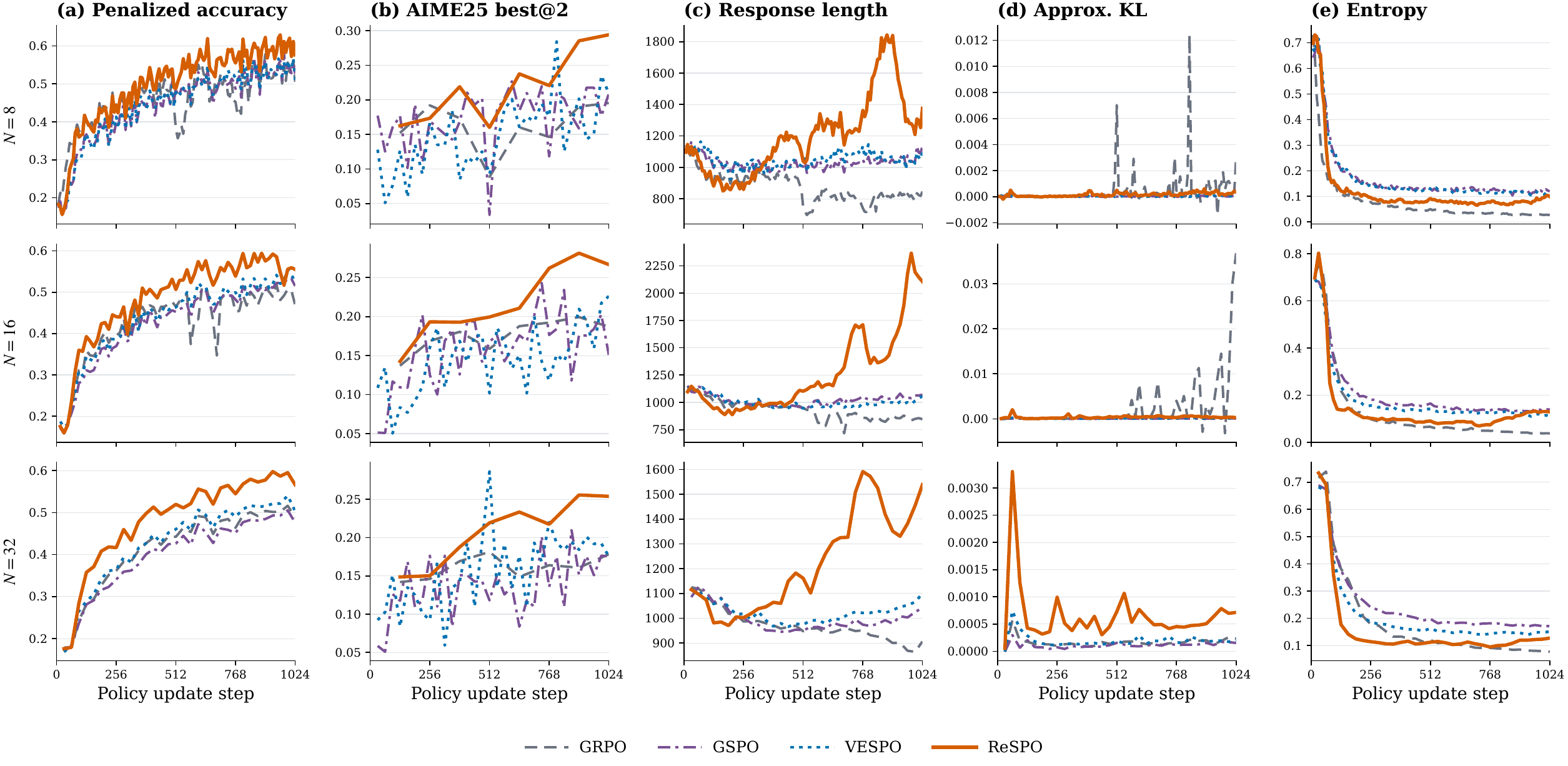}
\caption{Detailed training statistics for Qwen3-30B-A3B-Base on DAPO-MATH. The training panel reports penalized training accuracy, and the evaluation panel reports AIME25 best@2 accuracy, the largest common best-of-$k$ metric available for all four methods; the other panels follow Fig.~\ref{fig:stats_qwen17b_alpha1}.}
\label{fig:stats_qwen30b_alpha1}
\end{figure}

Fig.~\ref{fig:training_accuracy} uses the recorded \texttt{critic/score/mean} history. Each rollout batch supplies $N$ policy updates, so the horizontal coordinate is \texttt{training/global\_step}$\times N$. The plotting script truncates every curve at $1024$ policy updates, applies $(\texttt{score}+1)/2$ to the signed DAPO-MATH score, and validates the response cap and ReSPO negative-branch parameters. The early statistic is the maximum logged value through step $256$ and is reported without an error bar. For the late statistic, we linearly interpolate the curve over $[896,1024]$ and integrate its first and second moments to obtain the time-weighted mean and standard deviation. Figs.~\ref{fig:stats_qwen17b_alpha1} and~\ref{fig:stats_qwen30b_alpha1} use the same experiments and additionally report AIME25 evaluation accuracy, response length, approximate KL divergence, and entropy. For 1.7B at $N=32$, App.~\ref{app:w_diagnostics} resolves the aggregate response-length separation into positive-branch importance-weight and coefficient-mass statistics. The KL panel uses \texttt{actor/ppo\_kl} for GRPO/GSPO and \texttt{actor/approx\_kl} for VESPO/ReSPO.

\end{document}